\documentclass[11pt]{article}

\usepackage{acl_hy} 
\usepackage{times}
\usepackage{latexsym}

\usepackage[T1]{fontenc}
\usepackage[utf8]{inputenc}
\usepackage{microtype}
\usepackage{inconsolata}

\usepackage{graphicx}
\usepackage{amsmath} 

\usepackage{multirow}  
\usepackage{graphicx}  
\usepackage{booktabs}  
\usepackage{makecell}

\usepackage[table]{xcolor}

\usepackage{fdsymbol}

\usepackage[many]{tcolorbox}	

\usepackage{fontawesome} 
     
\tcbset{
    sharp corners,
    colback = white,
}   
\newtcolorbox{boxA}{
    width=\linewidth,
    left=5pt, right=5pt,
    boxsep=4pt,
    rounded corners,
    arc = 5pt 
}

\title{The Art of Efficient Reasoning: Data, Reward, and Optimization}

\author{
  Taiqiang Wu$^{\clubsuit\spadesuit}$\thanks{Equal contributions.}, Zenan Xu$^{\spadesuit}$\footnotemark[1], 
  Bo Zhou$^\spadesuit$$\thanks{Corresponding authors.}$, Ngai Wong$^\clubsuit$$^{\dag}$\\
  $^\clubsuit$The University of Hong Kong, $^\spadesuit$LLM Department, Tencent\hspace{0.3cm} \\
  \texttt{chaysezhou@tencent.com, nwong@eee.hku.hk} \\
\url{https://wutaiqiang.github.io/project/Art}
  }

\begin{document}
\maketitle
\thispagestyle{firstpage}%
\begin{abstract}
Large Language Models (LLMs) consistently benefit from scaled Chain-of-Thought (CoT) reasoning, but also suffer from heavy computational overhead.
To address this issue, efficient reasoning aims to incentivize short yet accurate thinking trajectories, typically through reward shaping with Reinforcement Learning (RL). 
In this paper, we systematically investigate the mechanics of efficient reasoning for LLMs.
For comprehensive evaluation, we advocate for more fine-grained metrics, including length distribution conditioned on correctness and performance across a wide spectrum of token budgets ranging from 2k to 32k.
First, we reveal that the training process follows a two-stage paradigm: \textit{length adaptation} and \textit{reasoning refinement}.
Through extensive experiments~(about 0.2 million GPU hours) in a unified protocol, we deconstruct training prompts and rollouts, reward shaping, and optimization strategies.
A central finding is to maintain a sufficient density of positive reward signals and avoid the short-is-correct trap.
Moreover, the learned length bias generalizes across domains and difficulty levels. 
We distill these findings into valuable insights and practical guidelines, and validate them across the Qwen3 models ranging from 0.6B to \textbf{30B}, demonstrating the robustness and generalization.
\end{abstract}

\section{Introduction}

Large language models (LLMs), such as Qwen3~\citep{yang2025qwen3} and DeepSeek-R1~\citep{guo2025deepseek}, have revolutionized the field of natural language processing~(NLP) due to their superior reasoning capabilities.
One key insight for such success is the consistently scaled Chain-of-thought~(CoT) thinking during inference~\citep{snell2024scaling}.
Despite the advancement, longer CoT also introduces significant trade-offs, such as high latency for real-world deployments~\citep{sui2025stop, wu2025revisiting}.

\begin{figure}[t]
  \includegraphics[width=\linewidth]{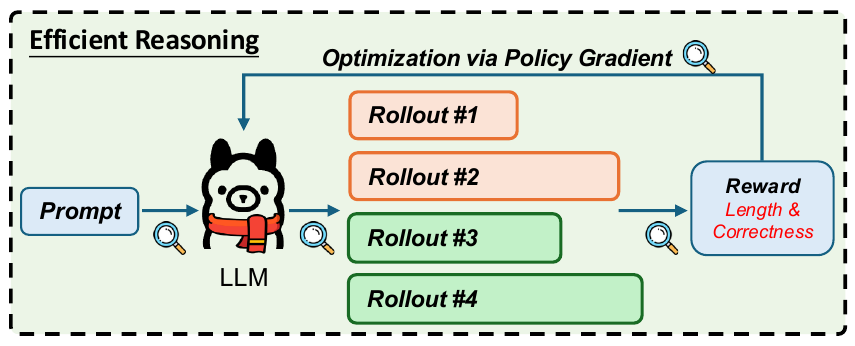}
  \caption{
  General pipeline for efficient reasoning via RL.
  The key is to promote short and accurate thinking trajectories via reward design.
  In this paper, we provide systematic insights (\raisebox{-0.25ex}{\includegraphics[height=0.8em]{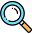}}) considering data, reward, and optimization.
  }
  \label{fig:efficient_framework}
\end{figure}

To address this issue, one mainstream method is to incentivize efficient reasoning via Reinforcement Learning~(RL) with reward shaping~\citep{ma2024highly, team2025kimi, liu2025learn}.
As shown in Figure \ref{fig:efficient_framework}, the core idea is to incentivize efficient reasoning by allocating the rewards based on rollout length and correctness.
For instance, one important principle is that shorter correct CoTs should receive higher rewards than longer correct CoTs~\citep{yeo2025demystifying}.
However, previous methods are almost exclusively focused on reward design~\citep{ hou2025thinkprune, liu2025learn}, while overlooking the broader training recipe, including data composition and optimization strategy.

\begin{figure*}[t]
  \includegraphics[width=\linewidth]{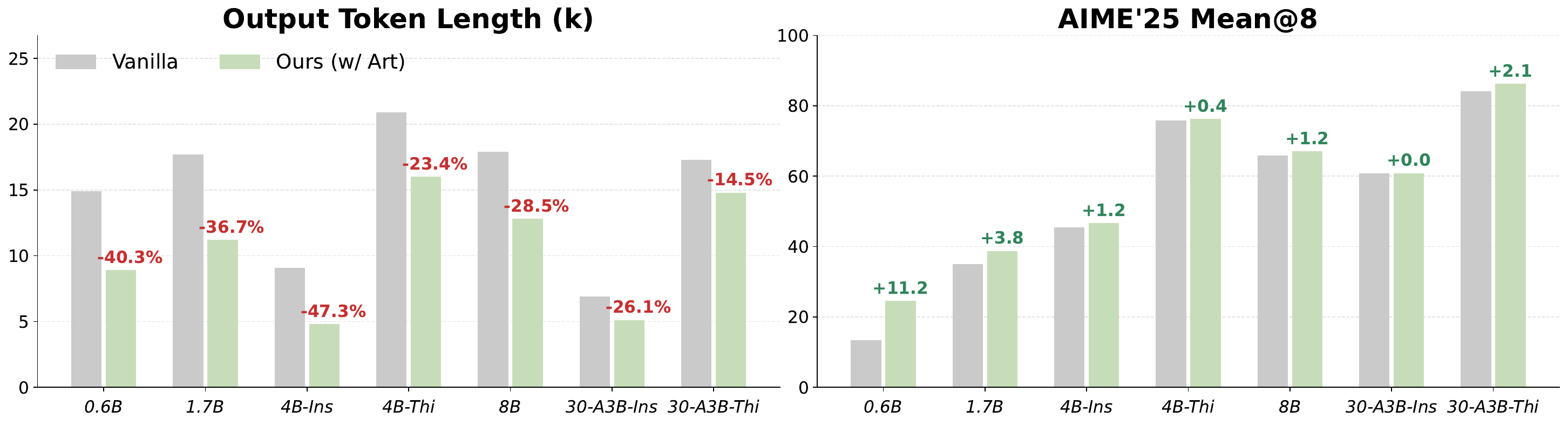}
  \caption{
  Performance comparison on AIME'25 for Qwen3 models ranging from 0.6B to 30B.
  }
  \label{fig:all_model_aime}
\end{figure*}

In this paper, we propose to \textit{systematically} investigate the mechanics of efficient reasoning in a unified experimental protocol.
Our analysis reveals that the training dynamics follow a two-stage paradigm, i.e., 1) \textit{length adaptation}, where the model rapidly adapts to token constraints; and 2) \textit{reasoning refinement}, where it optimizes performance within the length scope.
For comprehensive observations, we advocate for more fine-grained metrics.
Specifically, we propose to compare the length distribution conditioned on correctness for the training prompts.
Meanwhile, for the downstream benchmarks, we argue to record the performance across a wide spectrum of token budgets ranging from 2k to 32k.
The effectiveness of different strategies is budget-dependent, exhibiting distinct or even contradictory behaviors.
Importantly, we find that the \textbf{learned length bias can be generalized across domains}, i.e., training on mathematical prompts works well on the code task.

Through extensive ablation studies, we further deconstruct the impact of data difficulty, rollout number, reward, and optimization strategies.
Notably, we find that \textbf{training on relatively easier prompts} provides a denser positive reward signal, which is essential for stable reasoning distillation.
More rollouts contribute to better performance, but also bring heavier training costs.
For the reward assignment, we compare the strategy to assign a negative reward or mask corresponding rollouts.
Moreover, we further explore the off-policy strategy with different staleness to speed up the reasoning refinement stage.
We distill all the findings into valuable insights and practical guidelines, and evaluate them on the Qwen3 series.
Figure \ref{fig:all_model_aime} demonstrates that our strategy reduces the CoT length by 15\% to 47.3\% while preserving or improving Mean@8 and achieving favorable trade-offs overall.
In summary, our contributions are as follows:
\begin{itemize}
    \item We identify and characterize the two-stage paradigm for efficient reasoning, i.e., length adaptation followed by reasoning refinement.
    \item We introduce fine-grained metrics, providing a more comprehensive understanding of training dynamics.
    \item We provide a systematic exploration of the training recipe, offering practical insights into data, reward, and optimization that significantly improve the efficiency of CoT models.
\end{itemize}

\section{Preliminary}

\subsection{Experimental Setup}

\paragraph{Prompts.}
RL methods have been demonstrated as an effective way for reasoning.
Given an input prompt $x$ from a dataset $\mathcal{D}$, the LLM policy $\pi_\theta$ generates a set of $N$ reasoning trajectories (rollouts) $\{y_1, y_2, \dots, y_N\}$. 
The objective is to update $\pi_\theta$ using policy gradients derived from reward signals upon rollouts.
Data quality is critical for LLM reasoning~\citep{guo2025deepseek}.
Therefore, we employ the popular DeepScaleR as training prompts\footnote{\href{https://huggingface.co/datasets/hkust-nlp/Laser-Deepscaler-Dataset}{Dataset link}}.


\paragraph{Reward Engineering.}
In standard outcome-supervised RL, the reward function focuses solely on correctness. Let $\mathbb{I}(\cdot)$ denote the indicator function which returns 1 if the condition holds and 0 otherwise. 
The vanilla reward is defined as:
\begin{equation}
R_{\text{vanilla}}(x, y_i) = \mathbb{I}(y_i \text{ is correct}).
\end{equation}
To enforce efficient reasoning, we apply reward shaping to incentivize concise yet accurate rollouts. 
In this work, we select the truncation strategy as a baseline with
\begin{equation}
R_{T}(x, y_i) = \mathbb{I}(y_i \text{ is correct}) \cdot \mathbb{I}(L(y_i) \le L_{T}),
\end{equation}
where $L(y_i)$ denotes the token length of the $i$-th rollout, and $L_{T}$ represents the target length.
We first compare this approach against various baselines such as Kimi-1.5~\citep{team2025kimi} and Laser~\citep{liu2025learn}. 
Further details are provided in Appendix \ref{appendix: details_reward_shaping}.

\begin{figure*}[t]
    \centering
  \includegraphics[width=0.95\linewidth]{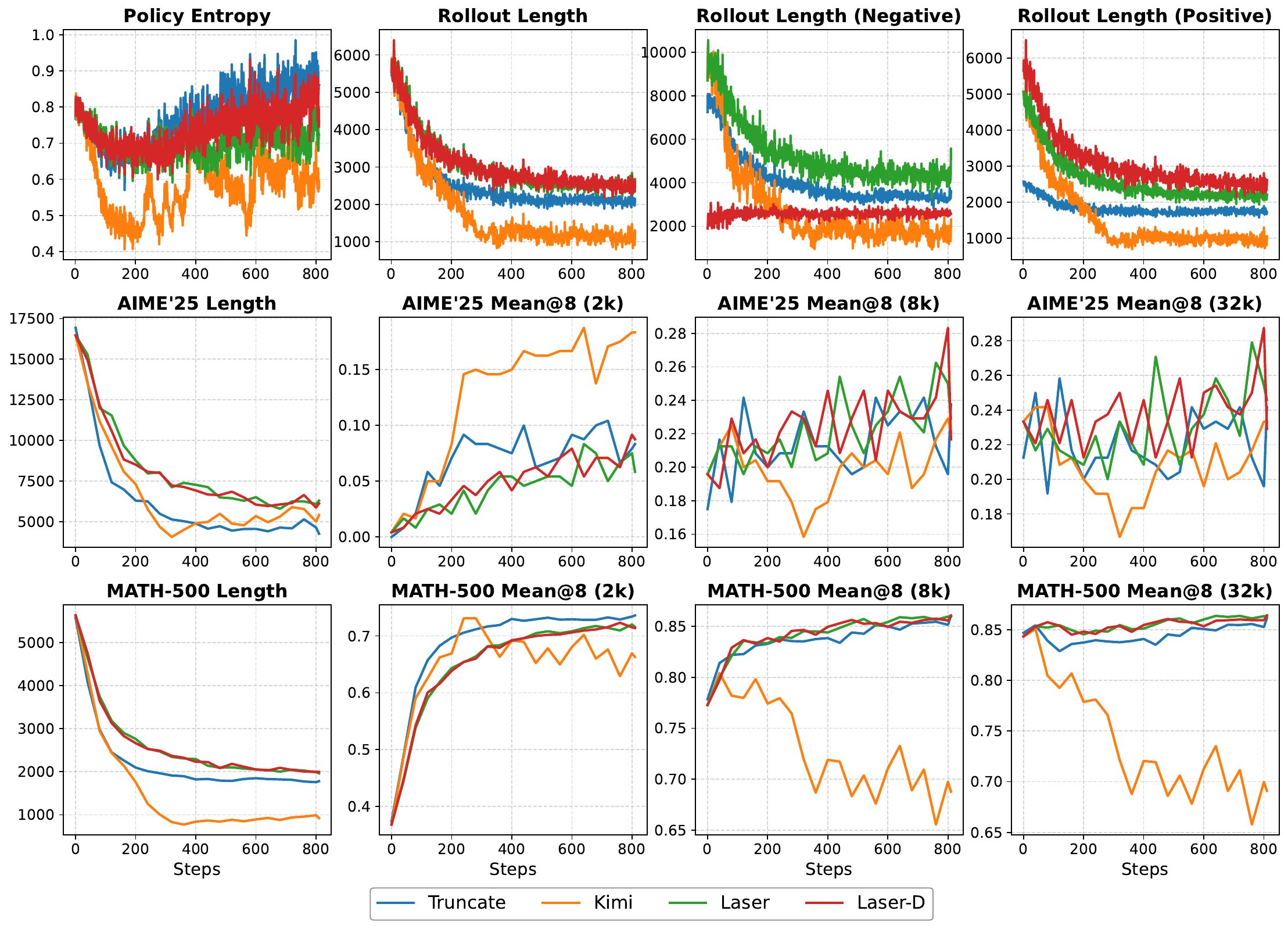}
  \caption{
  Training dynamics of various reward shaping methods on \texttt{DeepSeek-R1-Distill-Qwen-1.5B}.
  All of them follow the two-stage paradigm, including constraint satisfaction and optimization within a feasible region.
  The behaviors are distinct when evaluated under different token budgets.
  }
  \label{fig:two_stage}
\end{figure*}


\paragraph{Evaluation Protocol.}
To capture the nuances of efficient reasoning, we advocate for more fine-grained metrics as follows:
\begin{itemize}
\item \textbf{Training dynamics:} We monitor the length distribution conditioned on correctness to visualize how the model trades off verbosity for precision.
\item \textbf{Budget-aware benchmarking:} For downstream tasks, we report performance across a wide spectrum of inference token budgets ($\mathcal{B} \in \{2\text{k}, 4\text{k}, 8\text{k}, 16\text{k}, 32\text{k}\}$).
\end{itemize}
We report Pass@8 and Mean@8 metrics across standard mathematical reasoning benchmarks: AIME'25~\citep{AIME}, AMC~\citep{AMC}, MATH-500~\citep{hendrycks2021measuring}, Minerva Math~\citep{lewkowycz2022solving}, and Olympiad Bench~\citep{he2024olympiadbench}. Additionally, we assess code capabilities via LiveCodeBench (LCB)~\citep{jain2024livecodebench}.



\paragraph{Training Implementation.}
We use \texttt{DeepSeek-R1-Distill-Qwen-1.5B} as the backbone model. 
RL training is performed using Group Relative Policy Optimization (GRPO)~\citep{shao2024deepseekmath}.
The learning rate is set to $1 \times 10^{-6}$ with a clip-high ratio of 0.28 following \citet{yu2025dapo}.
During the rollout phase, we use a batch size of 128 and sample $N=8$ trajectories per prompt with a maximum length of 16k~($L_R=16k$). 
The target length $L_T$ is 4k.

\subsection{Two-stage Paradigm}


As illustrated in Figure \ref{fig:two_stage}, the training dynamics exhibit a two-stage paradigm:

\paragraph{Stage I: Length Adaptation.}
The optimization of constraint satisfaction dominates this initial phase.
Driven by the length penalty, the model rapidly adjusts its output distribution to avoid zero-reward truncation.
As shown in the \textit{Rollout Length} curves, the average token consumption undergoes a precipitous decline (e.g., from $\sim$6k to $\sim$2k), exhibiting an exponential decay pattern.
Simultaneously, the \textit{Policy Entropy} decreases significantly, indicating that the model is converging towards a subspace of shorter and valid trajectories.

\begin{figure*}[t]
  \includegraphics[width=0.95\linewidth]{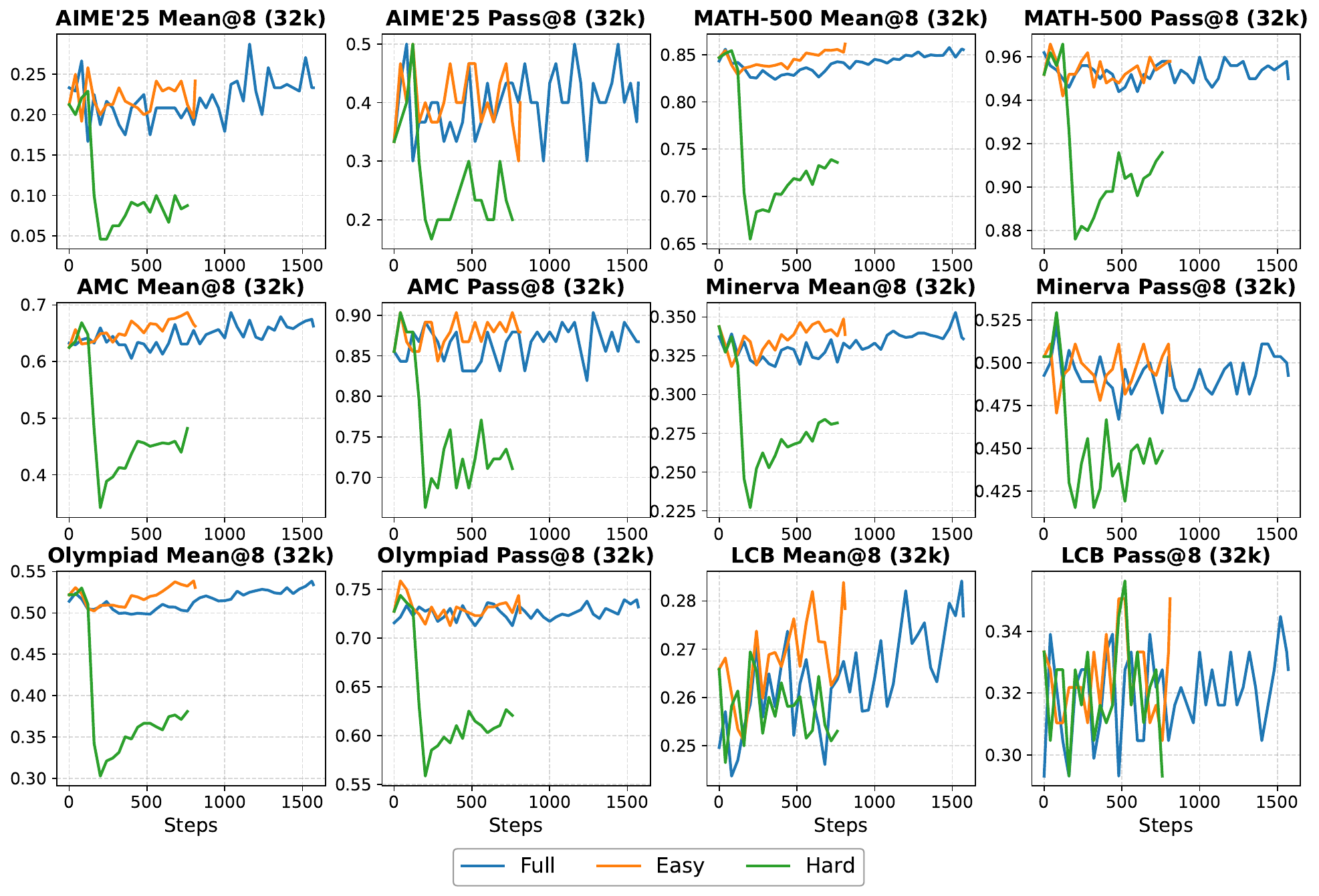}
  \caption{
  Performance training on all prompts and easy/hard counterparts~(rollout $L_R=16k$, target $L_T=4k$).
  }
  \label{fig:data_diff}
\end{figure*}

\paragraph{Stage II: Reasoning Refinement.}
Once the rollout length stabilizes within the target budget, the training enters a stationary phase regarding length, shifting focus to performance optimization.
In this stage, the length curves plateau, demonstrating that the model has successfully adapted to the hard constraints on output length.
Crucially, the performance metrics (e.g., Mean@8) continue to evolve or recover.
At the same time, the \textit{Policy Entropy} increases with such exploration.
This indicates that the model is learning to increase the information density of each token to improve accuracy without violating the length budget.

\subsection{Distinct Behaviors across Token Budgets}

Figure \ref{fig:two_stage} also reveals that model behaviors are highly sensitive to the token constraints, exhibiting distinct and even contradictory trends.

Under a strict budget (2k), performance is dominated by length adaptation. 
Aggressive penalties (e.g., Kimi) excel here by forcing the model to fit the narrow context window. 
However, under a generous budget (32k), such a strategy suffers from a reasoning collapse issue.
In contrast, Laser exhibits a U-shaped trajectory at 32k, which initially drops due to compression, but subsequently recovers through reasoning refinement. 
This decoupling phenomenon highlights a critical trade-off: over-optimizing for efficiency can severely harm the upper-bound reasoning capability, necessitating the multi-budget evaluation protocol~\citep{Qi2025OptimizingAR}.

\subsection{Generalization across Domains}

As detailed in Figures \ref{appendix:fig:two_stage_1} and \ref{appendix:fig:two_stage_2}, the dynamic on code task (i.e., LiveCodeBench) is similar to other mathematical tasks, though the training prompts are math-related only.
For example, on the LCB benchmark, the Kimi strategy performs best under a 2k budget but is comparable at larger budgets, which is similar to other benchmarks.
The similarity proves that learned length bias can be generalized across domains.
We further provide extensive results on private benchmarks in Appendix \ref{appendix:results_on_private}.

\section{Experiments and Guidelines}



\subsection{Data: Prompt and Rollout}


\paragraph{Impact of Prompt Difficulty.}
The difficulty of training prompts plays a pivotal role in determining the density of positive reward signals.
To investigate this, we split DeepScaleR prompts based on the pass rate over $N=8$ rollouts into DeepScaleR-Easy (pass rate $> 0.5$) and DeepScaleR-Hard (pass rate $\le 0.5$).


As illustrated in Figure \ref{fig:data_diff}, the training dynamics exhibit stark differences.
Training exclusively on hard prompts results in catastrophic failure. 
The policy entropy spikes drastically, and the rollout length collapses prematurely. 
Consequently, downstream performance metrics (e.g., Mean@8 on AMC and Olympiad Bench) degrade significantly. 
This suggests that when the model struggles to generate correct answers, the RL signal becomes \textbf{dominated by the length penalty on incorrect rollouts, leading to reasoning collapse}.
We attribute such an issue to the \textbf{sparsity of positive samples}, which leads to the overfitting on short output length.
Conversely, training on the easier counterpart yields the most stable trajectory. 
The policy entropy remains low and stable, indicating consistent positive reinforcement. 
The rollout length adapts smoothly to the target budget.
Crucially, despite training on easy prompts, the performance on relatively tough tasks~(e.g., AIME'25) is comparable to (or even slightly exceeding) training on the full dataset.

Please refer to Appendix \ref{appendix: details_data_select} for results and analysis under more settings.
Appendix \ref{appendix:cases} further indicates cases for reasoning collapse from baselines and paradigm shift from our strategy.

\begin{figure}[t]
  \includegraphics[width=\linewidth]{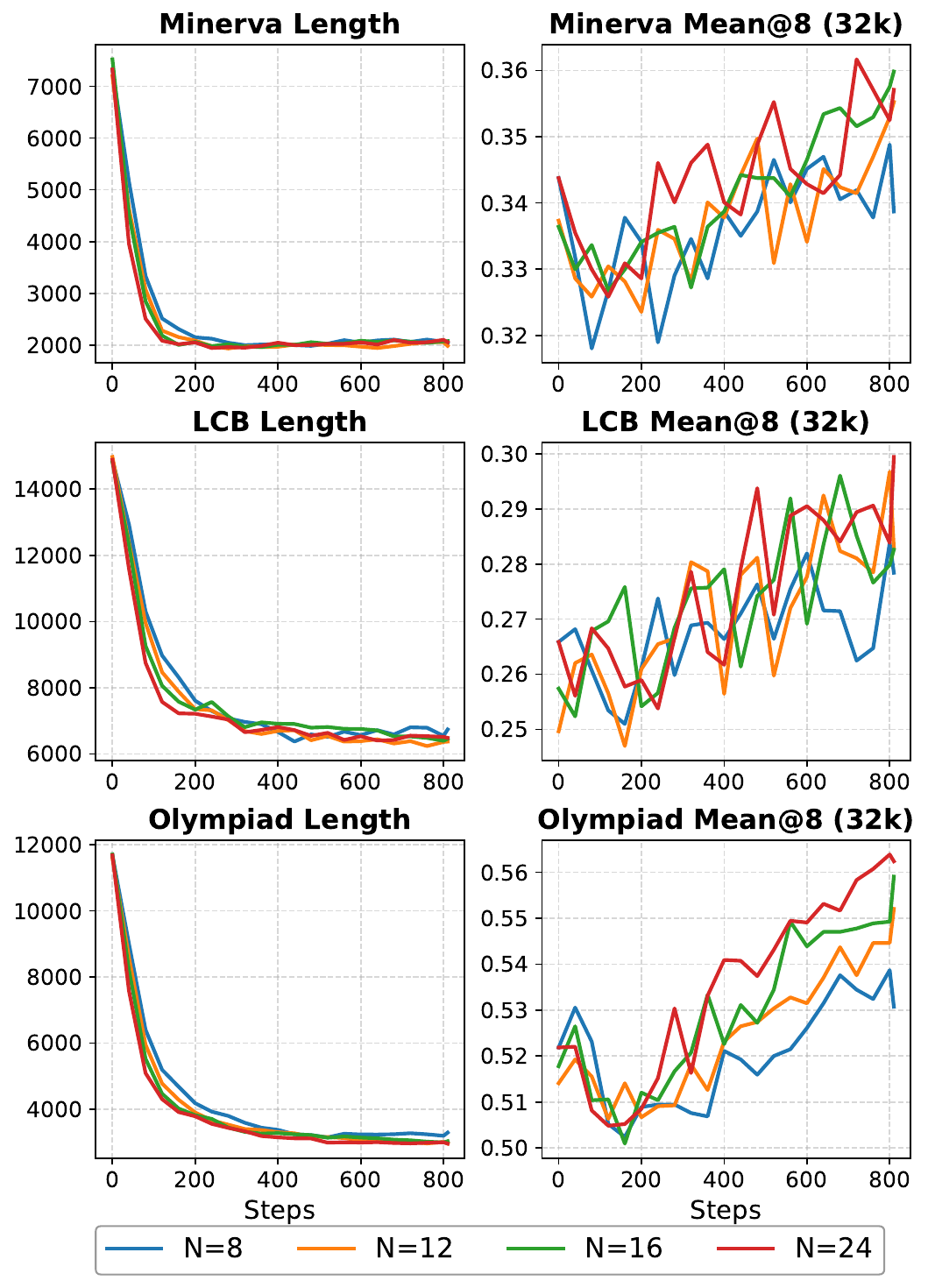}
  \caption{
  Performance with various rollouts $N$ using DeepScaleR-Easy.
  }
  \label{fig:roll_diff}
\end{figure}

\paragraph{Impact of Rollout Number $N$.}



Given that the density of valid reward signals is crucial, we further investigate the impact of the rollout number $N$. 
We set $N \in \{8, 12, 16, 24\}$ with a rollout limit $L_R=16k$ and a target length $L_T=4k$ on the DeepScaleR-Easy prompts.

As shown in Figure \ref{fig:roll_diff}, increasing $N$ yields observable benefits that significantly speed up the \textit{Length Adaptation} phase.
With a larger N, it is easier to discover short and correct trajectories, promoting the length curve (e.g., $N=24$, red line) to decay faster than the baseline ($N=8$, blue line), though all settings converge to a similar length floor.
Meanwhile, larger $N$ leads to a more robust \textit{Reasoning Refinement} stage.
In mathematical benchmarks, the model recovers its reasoning capabilities faster and achieves a higher asymptotic Mean@8.

However, this advantage is task-dependent. 
On the LiveCodeBench (LCB) coding task, the performance gap between $N=8$ and $N=24$ is marginal, suggesting that the complexity of code generation may require distinct exploration strategies beyond simply scaling $N$.
Please refer to Appendix \ref{appendix: details_rollout_N} for results under more settings.

\begin{boxA}
\textbf{Insights towards Training Data}:
\textit{
The key is to ensure sufficient and effective rewards.
Training on easier prompts allows LLMs to focus on length reduction without compromising performance.
Larger rollout $N$ would be better if computational resources allow.
}
\end{boxA}

\subsection{Reward on Negative Rollouts}

\begin{table}[h]
\centering

\begin{tabular}{l c c c c}
\toprule
\multirow{2}{*}{\textbf{Strategy}} & \multicolumn{2}{c}{\textbf{Correct}} & \multicolumn{2}{c}{\textbf{Incorrect}} \\
\cmidrule(lr){2-3} \cmidrule(lr){4-5}
 & \textbf{Short} & \textbf{Long} & \textbf{Short} & \textbf{Long} \\
\midrule
Vanilla      & 1 & 0 & 0 & 0 \\
-I           & 1 & 0 & - & - \\
-L\&C        & 1 & - & 0 & 0 \\
-L\&C-S\&I   & 1 & - & - & 0 \\
-L\&C-L\&I   & 1 & - & 0 & - \\
\bottomrule
\end{tabular}

\caption{Reward for different strategies on negative rollouts.
$-$ denotes masking out.}
\label{tab:reward_strategy}
\end{table}



The art of RL is to utilize the negative signals.
In the standard truncation strategy~(denoted as Vanilla), both \textit{incorrect} responses and \textit{overlong correct} responses are treated as negative samples ($R=0$).
An alternative way is to mask these negative samples rather than setting the reward to 0.
To investigate this, we conduct a fine-grained ablation study by masking specific subsets of negative rollouts, as detailed in Table \ref{tab:reward_strategy}.

\begin{figure*}[t]
  \includegraphics[width=\linewidth]{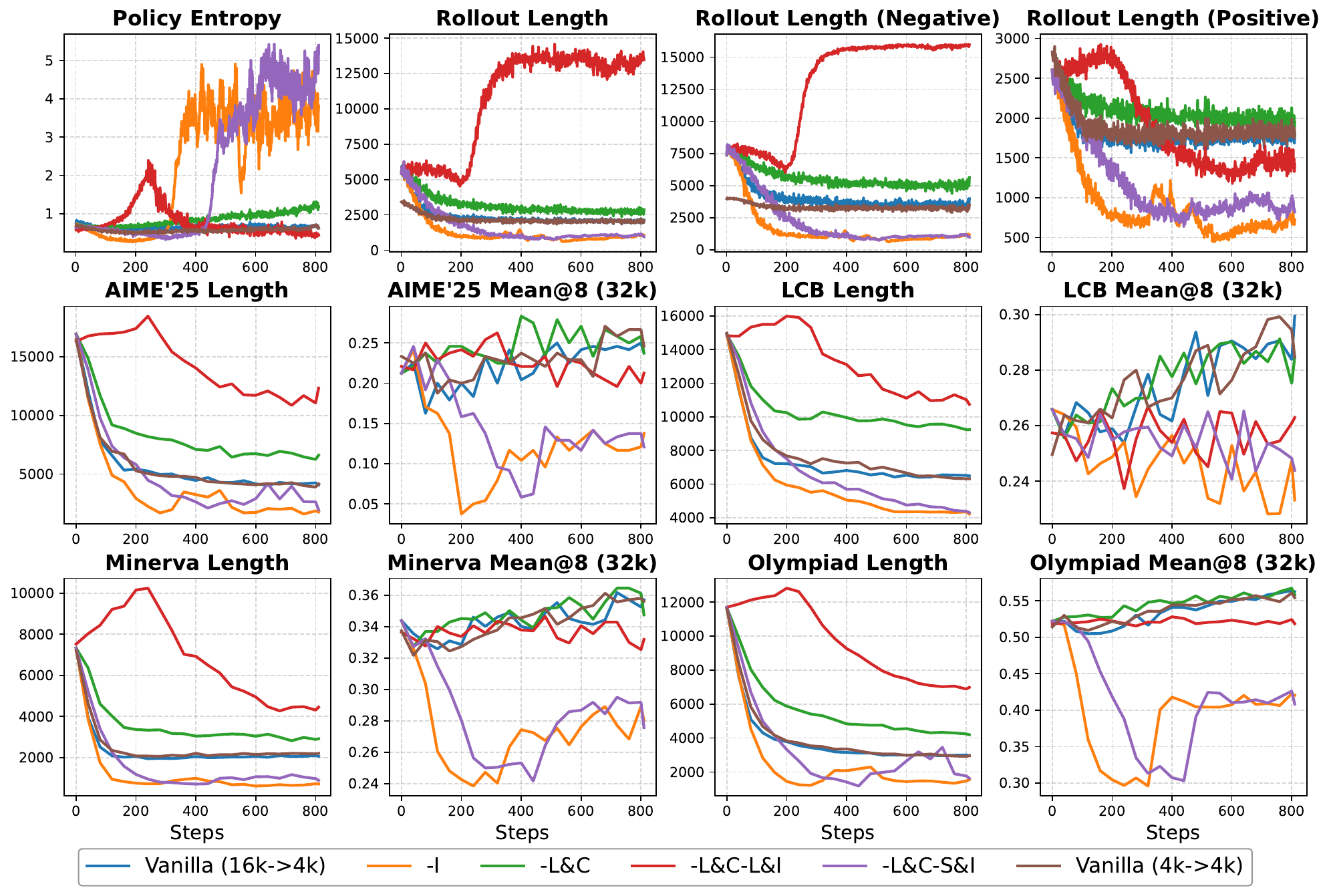}
  \caption{
  Performance for various reward strategies on negative rollouts (rollout $L_R=16k$, target $L_T=4k$, $N=24$).
  We also visualize $L_R=4k$, $L_T=4k$ for comparison.
  }
  \label{fig:reward_neg}
\end{figure*}

As illustrated in Figure \ref{fig:reward_neg}, the training dynamics reveal that different masking strategies exhibit distinct behaviors.
In short, improper strategies will lead to unintended consequences, categorized into three distinct failure modes:

\paragraph{1) The trap that short is correct (-I, -L\&C-S\&I).}
When the correctness is coupled with length, the model will be misled by such incorrect causal relationships.
For masking all incorrect rollouts (-I), we only penalize overlong and correct rollouts.
The training signal only contains a) short and correct rollouts with positive reward and b) overlong and correct rollouts with negative reward. 
In this way, the model would hack this bias to generate short output.
As shown by the \textcolor{orange}{orange line}, the policy entropy explodes after 400 steps and rollout length collapses precipitously. 
The model abandons reasoning entirely to satisfy the length constraint.
Meanwhile, when masking overlong correct and short incorrect rollouts~(-L\&C-S\&I), the dynamics are similar.
The training signal only contains a) short and correct rollouts with positive reward and b) overlong and incorrect rollouts with negative reward. 

\paragraph{2) Short rollouts only.}
A particularly interesting phenomenon occurs in the strategy \textbf{-L\&C-L\&I} (\textcolor{red}{red line}), where we mask all overlong trajectories.
It means that the LLMs are optimized with short correct and short incorrect rollouts exclusively.
In this setting, the overlong outputs are masked, without either positive or negative rewards.
After 200 steps (reasoning refinement stage), the LLMs hack this and begin to generate overlong outputs.
Interestingly, these outputs are almost incorrect, but the models \textit{do not} collapse.

\paragraph{3) Do not penalize overlong but correct rollouts (-L\&C).}
The \textbf{-L\&C} strategy (\textcolor{green}{green line}) masks overlong but correct rollouts instead of penalizing them.
On the downstream benchmarks, the LLMs will generate longer outputs and also outperform the vanilla baseline, indicating a trade-off for length control and performance.

\begin{figure*}[!t]
  \includegraphics[width=0.95\linewidth]{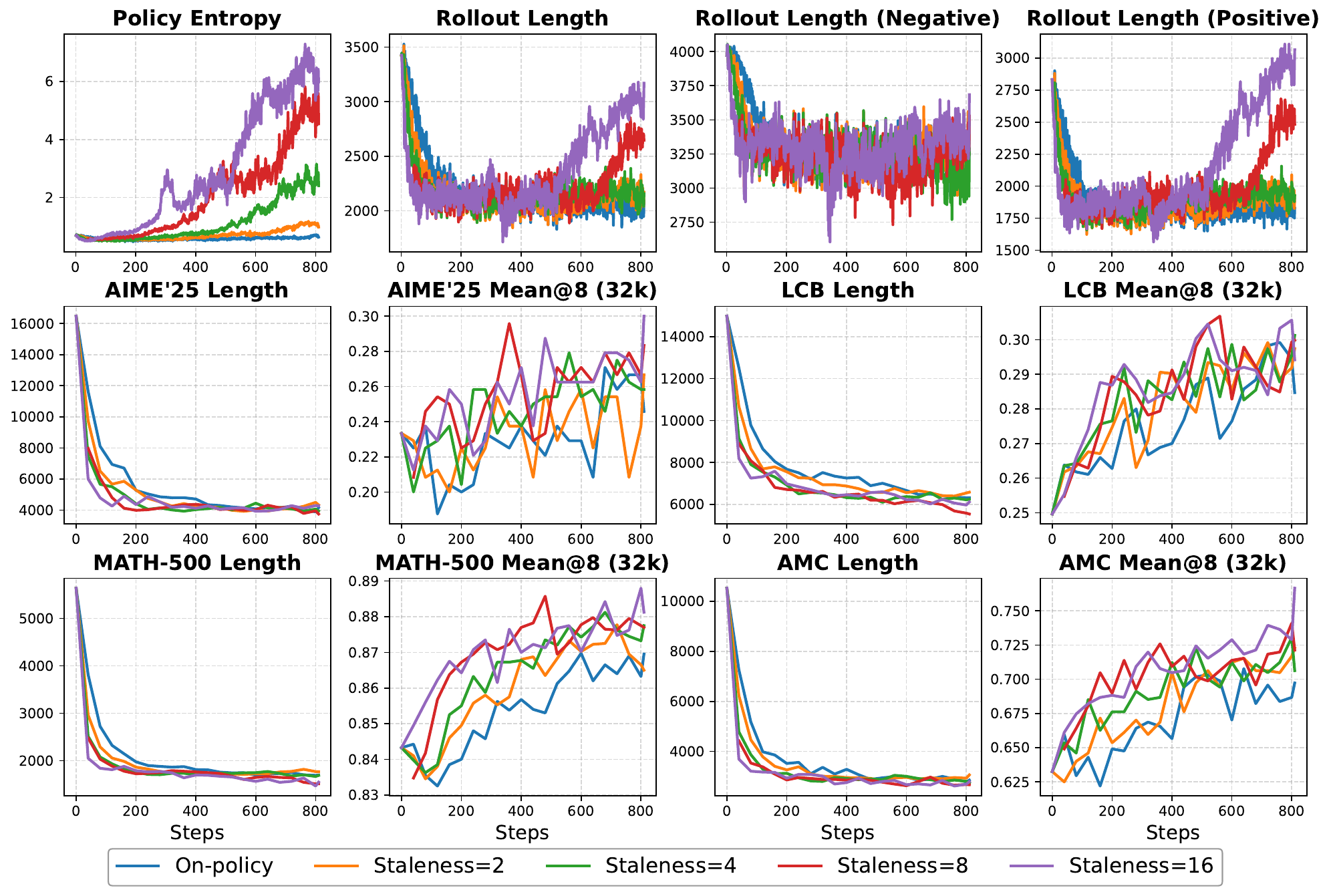}
  \caption{
  Performance for off-policy strategy with various staleness~(i.e., 2,4,8,16).
  }
  \label{fig:off_policy}
\end{figure*}

\paragraph{Length of Negative Samples.}
Additionally, we compare these complex shaping strategies against a simple baseline sampling at target length (i.e., $L_R=L_T=4k$, \textcolor{brown}{brown line}).
Compared to Vanilla~($L_R=16k, L_T=4k$), the positive samples are roughly the same (correct rollouts that are less than $4k$) while the negative samples are much shorter (4k vs. 6k).
We can observe that it achieves the optimal Pareto frontier.
We attribute such success to avoiding the harmful explicit length bias trap that short is correct.
Typically, the positive rollouts are shorter than the negative ones, which implicitly encourages the model to be short yet accurate.

\begin{boxA}
\textbf{Insights towards Reward on Negative Rollouts}:
\textit{
Not penalizing overlong correct rollouts leads to higher performance, but also slightly longer outputs.
Sampling at target length ($L_R=L_T$) achieves a better trade-off via avoiding the length trap.
}
\end{boxA}

\subsection{Off-policy Optimization}


Off-policy optimization via introducing staleness can significantly accelerate the training process but also introduces instability~\citep{zheng2025prosperity, huang2025low}.
Meanwhile, we can observe that the performance keeps growing at the reasoning refinement stage.
Hence, we conduct experiments with varying staleness degrees $S \in \{2, 4, 8, 16\}$ using the robust setting derived from our previous guidelines ($L_R=L_T=4k, N=24$).




As illustrated in Figure \ref{fig:off_policy}, the results reveal a trade-off between performance and stability.
Firstly, the \textit{Length Adaptation} stage is significantly shortened. 
As staleness increases, the rollout length decays more rapidly, allowing the model to satisfy the token budget earlier.
After that, the model enters the \textit{Reasoning Refinement} stage and learns quickly.
As a result, performance on downstream tasks shows that high-staleness models (e.g., $S=16$) achieve higher accuracy than the on-policy baseline.


Contrary to findings that excessive off-policy updates lead to catastrophic model collapse, our experiments maintain high performance. 
We attribute this robustness to our optimized setup (easy prompts and large $N$), which ensures sufficient and effective reward signals.
However, larger staleness also introduces \textit{potential instability risks}:
i) \textit{Entropy explosion:} 
As shown in the \textit{Policy Entropy} subplot, high staleness (e.g., 16) causes a dramatic surge in entropy after 400 steps.
For the on-policy baseline, the entropy remains almost the same as initialization.
ii) \textit{Length rebound:} 
The positive rollout length on training prompts under high staleness begins to drift upwards again, indicating that the policy model is struggling to balance efficiency and effectiveness.

\begin{boxA}
\textbf{Insights towards Off-policy Optimization}:
\textit{
Appropriate staleness can accelerate convergence without harming accuracy (before 800 steps), but also introduces latent instability, such as rising entropy.
We suggest the on-policy strategy, especially for larger and more fragile LLMs.
}
\end{boxA}

\section{Extensive Analysis}

\subsection{Evaluated on More LLMs}


To verify the universality of our derived guidelines, we extend our evaluation to the Qwen3 family, ranging from 0.6B to 30B.
Please refer to Appendix \ref{appendix:qwen3_settings} for detailed hyperparameters.
After training, we evaluate these compressed LLMs on AIME'25~\citep{AIME}.

As shown in Table \ref{tab:more_llms}, the experimental results demonstrate the robustness of our strategy across different model scales.
Specifically, on the \texttt{Qwen3-0.6B}, our method significantly boosts the Mean@8 from 13.33 to 24.58, while reducing the average response length from 14.9k to 8.9k.
Meanwhile, our approach improves the Pass@8 to 36.67 compared to the vanilla score of 26.67. 
A similar trend is observed in larger models.
For instance, on \texttt{Qwen3-4B-Instruct-2507}, our method maintains a superior Pass@8 of 70.00 and increases the Mean@8 from 45.42 to 46.67, while drastically compressing the length from 9.1k to roughly 4.8k.

\subsection{Case Study}

To qualitatively validate the impact of our strategy, we compare the reasoning trajectories of the vanilla and optimized models.
Please refer to Appendix \ref{appendix:cases} for detailed examples and analysis.
In short, our method not only encourages short outputs but also incentivizes the model to reorganize its CoTs into a more streamlined and expert-like format.

\begin{table}[!t]
\centering
\resizebox{\linewidth}{!}{
\begin{tabular}{l c c c}
\toprule
\textbf{Method} & \textbf{Mean@8}$\uparrow$ & \textbf{Pass@8}$\uparrow$ & \textbf{Length}$\downarrow$ \\
\midrule
\multicolumn{4}{c}{\texttt{Qwen3-0.6B}} \\
\cmidrule(r){1-4}
Vanilla & 13.33 & 26.67 & 14.9k \\
\rowcolor[RGB]{235,245,231} Ours (step 640) & 24.58 & 36.67 & 8.9k \\
\midrule
\multicolumn{4}{c}{\texttt{Qwen3-1.7B}} \\
\cmidrule(r){1-4}
Vanilla & 35.00 & 60.00 & 17.7k \\
\rowcolor[RGB]{235,245,231} Ours (step 560) & 38.75 & 60.00 & 11.2k \\
\midrule
\multicolumn{4}{c}{\texttt{Qwen3-4B-Instruct-2507}} \\
\cmidrule(r){1-4}
Vanilla & 45.42 & 66.67 & 9.1k \\
\rowcolor[RGB]{235,245,231} Ours (step 1440) & 46.67 & 70.00 & 4.8k
\\
\midrule
\multicolumn{4}{c}{\texttt{Qwen3-4B-Thinking-2507}} \\
\cmidrule(r){1-4}
Vanilla & 75.83 & 90.00 & 20.9k \\
\rowcolor[RGB]{235,245,231} Ours (step 200) & 76.25 & 86.67 & 16.0k
\\
\midrule
\multicolumn{4}{c}{\texttt{Qwen3-8B}} \\
\cmidrule(r){1-4}
Vanilla & 65.83 & 86.67 & 17.9k \\
\rowcolor[RGB]{235,245,231} Ours (step 100) & 67.08 & 83.33 & 12.8k
\\
\midrule
\multicolumn{4}{c}{\texttt{Qwen3-30B-A3B-Instruct-2507}} \\
\cmidrule(r){1-4}
Vanilla & 60.83 & 83.33 & 6.9k \\
\rowcolor[RGB]{235,245,231} Ours (step 600) & 60.83 & 76.67 & 5.1k
\\
\midrule
\multicolumn{4}{c}{\texttt{Qwen3-30B-A3B-Thinking-2507}} \\
\cmidrule(r){1-4}
Vanilla & 84.17 & 96.67 & 17.3k \\
\rowcolor[RGB]{235,245,231} Ours (step 120) & 86.25 & 96.67 & 14.8k
\\
\bottomrule
\end{tabular}
}
\caption{Performance on AIME'25 for Qwen3 models.}
\label{tab:more_llms}
\end{table}

\section{Related Work}

\subsection{Efficient Reasoning}


Efficient reasoning methods aim to mitigate the overthinking phenomenon~\citep{wu2025more, sui2025stop} and reduce the prohibitive inference costs associated with long-form CoT~\citep{wu2025revisiting, cui2024survey}. 
One prominent approach trains long CoTs to be short using SFT~\citep{xia2025tokenskip, ma2025cot} or RL~\citep{hou2025thinkprune, shen2025dast, liu2025learn, liang2025deepcompress}. 
Parallel research directions explore architectural innovations, such as reasoning within latent spaces~\citep{hao2024training, su2025token} or more efficient decoding~\citep{sun2024fast, xu2025phi}.
We refer the readers to \citet{feng2025efficient} for more details.
Unlike works that propose novel architectures, we focus on the mechanics of \textit{RL-based} efficiency optimization.

\subsection{Reward Shaping Methods}

The philosophy for reward shaping is to incentivize short yet accurate rollouts via allocating training rewards~\citep{weng2025think}.
The first principle is to promote shorter responses and penalize longer responses among correct ones~\citep{team2025kimi, hou2025thinkprune, aggarwal2025l1}.
Meanwhile, we can also penalize longer for incorrect answers~\citep{team2025kimi}.
Contrary to that, \citet{liu2025learn, yeo2025demystifying} argue to promote longer incorrect rollouts to encourage exploration.
Despite these advancements, existing studies often evaluate reward functions in isolation.
In this paper, we select the simplest truncation strategy and conduct extensive experiments in a unified protocol.
\section{Conclusion}

In this work, we first reveal that the training dynamics of efficient reasoning follow a two-stage paradigm.
Meanwhile, we advocate for more fine-grained metrics.
Based on these, we further deconstruct the impact of data difficulty, rollout number, reward on negative rollouts, and optimization strategies.
All findings are distilled into valuable insights and practical guidelines.
A key finding is to train on relatively easier prompts for sufficient and effective rewards.
The learned length bias for reasoning can be generalized across domains and difficulty levels.
Evaluation of more LLMs up to 30B further demonstrates the robustness and generalization.


\section*{Limitation and Future Work}

In this work, we systematically investigate the mechanics of efficient reasoning. 
However, there are several limitations for future research.

\paragraph{Domain diversity.}
In this paper, we train on DeepScaleR, containing mathematical reasoning prompts, and validate on math and coding benchmarks.
One future work is to evaluate on more domains, such as creative writing.
Another interesting topic is whether training on more diverse prompts contributes to better performance.


\paragraph{Adaptive length.}
In this paper, we employ fixed rollout and target length~(i.e., $L_R$ and $L_T$).
One intuitive idea is to set the length adaptively based on the prompts and current LLMs.
For target length $L_T$, we conduct experiments on a toy setting detailed in Appendix \ref{appendix:adaptive_l_t}.
For adaptive rollout length $L_R$, we can pre-compute and set the rollout engine correspondingly.

\paragraph{Evaluate on larger models.}
In this paper, we conduct extensive experiments (about 0.2 million GPU hours) in a unified protocol on the \texttt{DeepSeek-R1-Distill-Qwen-1.5B}.
Moreover, we extend our evaluation to the Qwen3 family, such as \texttt{Qwen3-30B-A3B-Instruct-2507}.
However, due to the limited GPUs, we do not evaluate on extremely large LLMs such as \texttt{Qwen3-235B-A22B-Instruct-2507}.
We leave it for future work.

\paragraph{More fine-grained supervision.}
In this paper, we focus on efficient reasoning via reward shaping-based RL training.
The principle is to incentivize short yet
accurate thinking trajectories.
We do not apply fine-grained refinement to the CoTs.
Meanwhile, we humans simplify the previous experience by making and employing useful tools, such as a notebook and a calculator.
Therefore, how to create useful tools and reuse them during reasoning remains a valuable topic.

\bibliography{ref}
\appendix
\section*{Appendix}

\section{Detailed Results on Reward Engineering}
\label{appendix: details_reward_shaping}



For Kimi-1.5~\citep{team2025kimi}, they first define a normalized length term:
\begin{equation}
    \tilde{L}(y_i) = \frac{L(y_i)-L_{\min}}{L_{\max}-L_{\min}},
\end{equation}
where $L_{\max}$ and $L_{\min}$ are the maximum and minimum length within rollout group.
Then the reward function can be rewritten as the sum of the correct reward and the incorrect penalty:
\begin{equation}
\small
\begin{aligned}
R_{\text{Kimi}}(x, y_i) = \mathbb{I}(y_i \text{ is correct}) \cdot \left(1+\alpha(0.5 - \tilde{L}(y_i)) \right) \\
+ \mathbb{I}(y_i \text{ is incorrect}) \cdot \min\left(0,\alpha(0.5 - \tilde{L}(y_i))\right).
\end{aligned}
\end{equation}

For Laser~\citep{liu2025learn}, it can be viewed as a base reward for correctness plus a bonus if the length condition is met:
\begin{equation}
\small
R_{\text{Laser}}(x, y_i) = \mathbb{I}(y_i \text{ is correct}) \cdot \left( 1 + \alpha \cdot \mathbb{I}(L(y_i) < L_T) \right).
\end{equation}

They also propose a variant to encourage exploration by adding an extra term to incentivize longer trajectories.
Specifically, it adds an exploration bonus for incorrect rollouts:
\begin{equation}
\small
\begin{aligned}R_{\text{Laser-D}}(x, y_i) &= \mathbb{I}(y_i \text{ is correct}) \cdot \left( 1 + \alpha \cdot \mathbb{I}(L(y_i) < L_T) \right) \\ &+ \mathbb{I}(y_i \text{ is incorrect}) \cdot \left( \alpha \cdot \mathbb{I}(L(y_i) \ge L_T) \right).
\end{aligned}
\end{equation}

\begin{figure*}[t]
  \includegraphics[width=\linewidth]{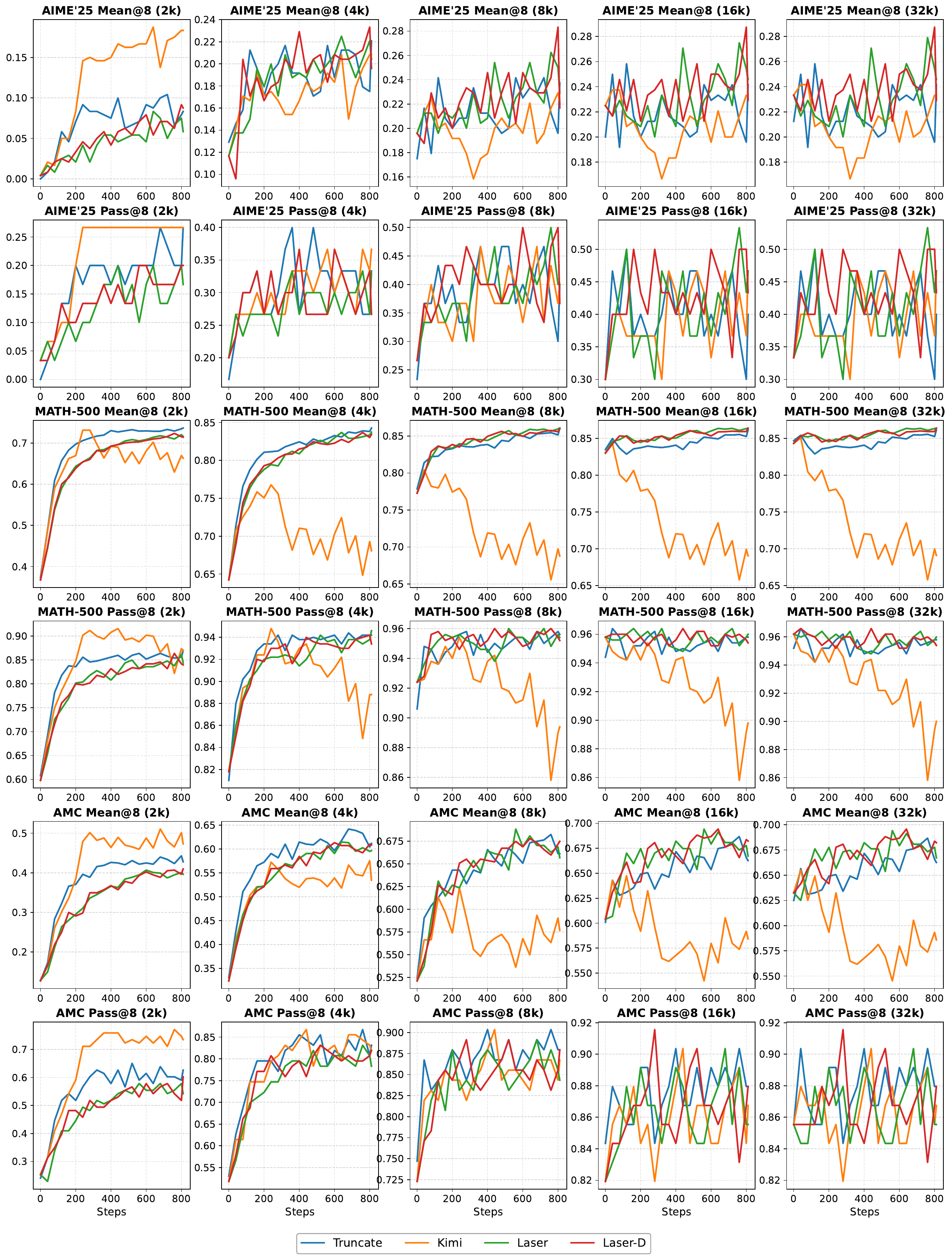}
  \caption{
  Training dynamics of various reward shaping methods on AIME'25, MATH-500, and AMC.
  }
  \label{appendix:fig:two_stage_1}
\end{figure*}
\begin{figure*}[t]
  \includegraphics[width=\linewidth]{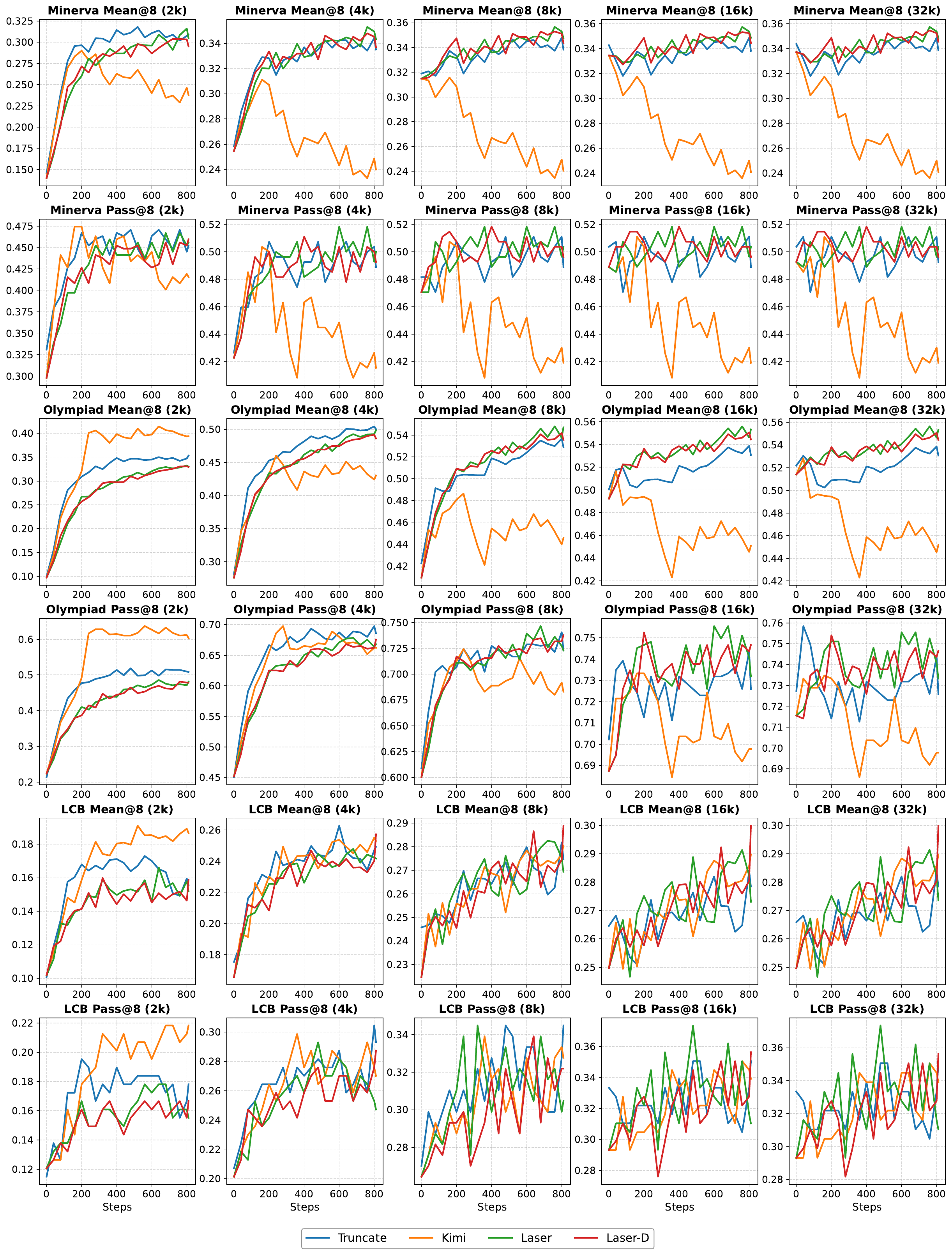}
  \caption{
  Training dynamics of various reward shaping methods on Minerva Math, Olympiad Bench, and LiveCodeBench.
  }
  \label{appendix:fig:two_stage_2}
\end{figure*}

In this paper, we set $\alpha$ to 0.4.
Figures \ref{appendix:fig:two_stage_1} and \ref{appendix:fig:two_stage_2} present the training dynamics on additional benchmarks, including AIME'25, MATH-500, AMC (Figure \ref{appendix:fig:two_stage_1}), and Minerva Math, Olympiad Bench, LiveCodeBench (Figure \ref{appendix:fig:two_stage_2}).

We can find that the training process strictly follows a two-stage paradigm, i.e., the \textit{length adaptation} phase followed by the \textit{reasoning refinement} phase.
The rollout length (top row) consistently decays before stabilizing. 
Meanwhile, the performance shows distinct behaviors across various token budgets.
For instance, a more aggressive strategy Kimi often exhibits gains at stricter budgets~(2k), but suffers from stagnation or collapse at generous budgets~(32k), whereas the truncation baseline maintains a more balanced recovery.
Without results under larger token budgets, the observations and conclusions will be biased.
These extensive results reinforce the necessity of our proposed fine-grained evaluation protocol.

\section{Detailed Results for Data Selection}
\label{appendix: details_data_select}


In the main paper, we demonstrate that training on easy prompts allows LLMs to focus on length reduction without compromising performance, which is attributed to sufficient and effective rewards.
Here, we provide details on all benchmarks and more settings.

\begin{figure*}[t]
  \includegraphics[width=\linewidth]{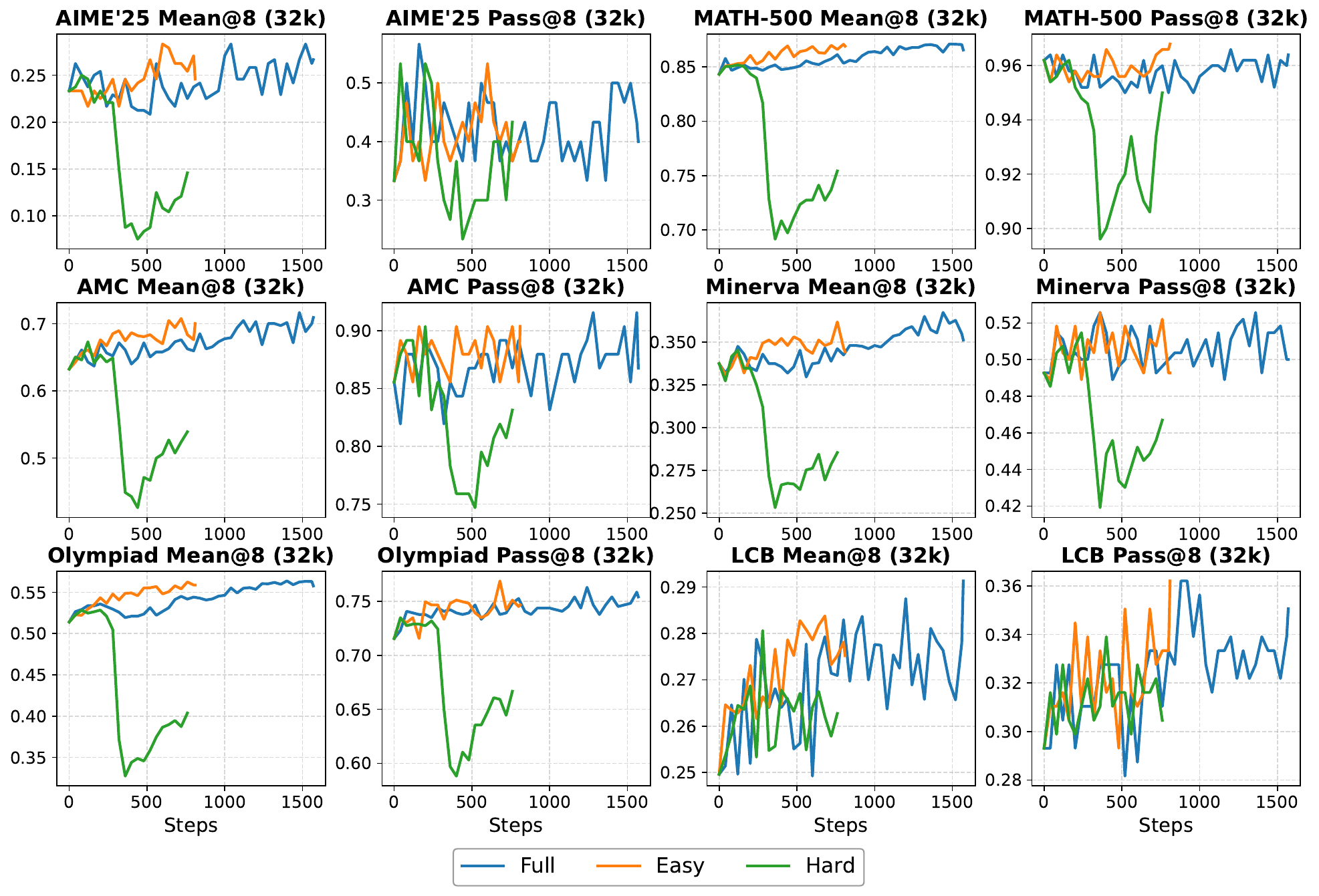}
  \caption{
  Performance training on various training prompts (rollout $L_R=16k$, target $L_T=8k$).
  }
  \label{appdix:fig:data_diff_16k_8k}
\end{figure*}

\begin{figure*}[t]
  \includegraphics[width=\linewidth]{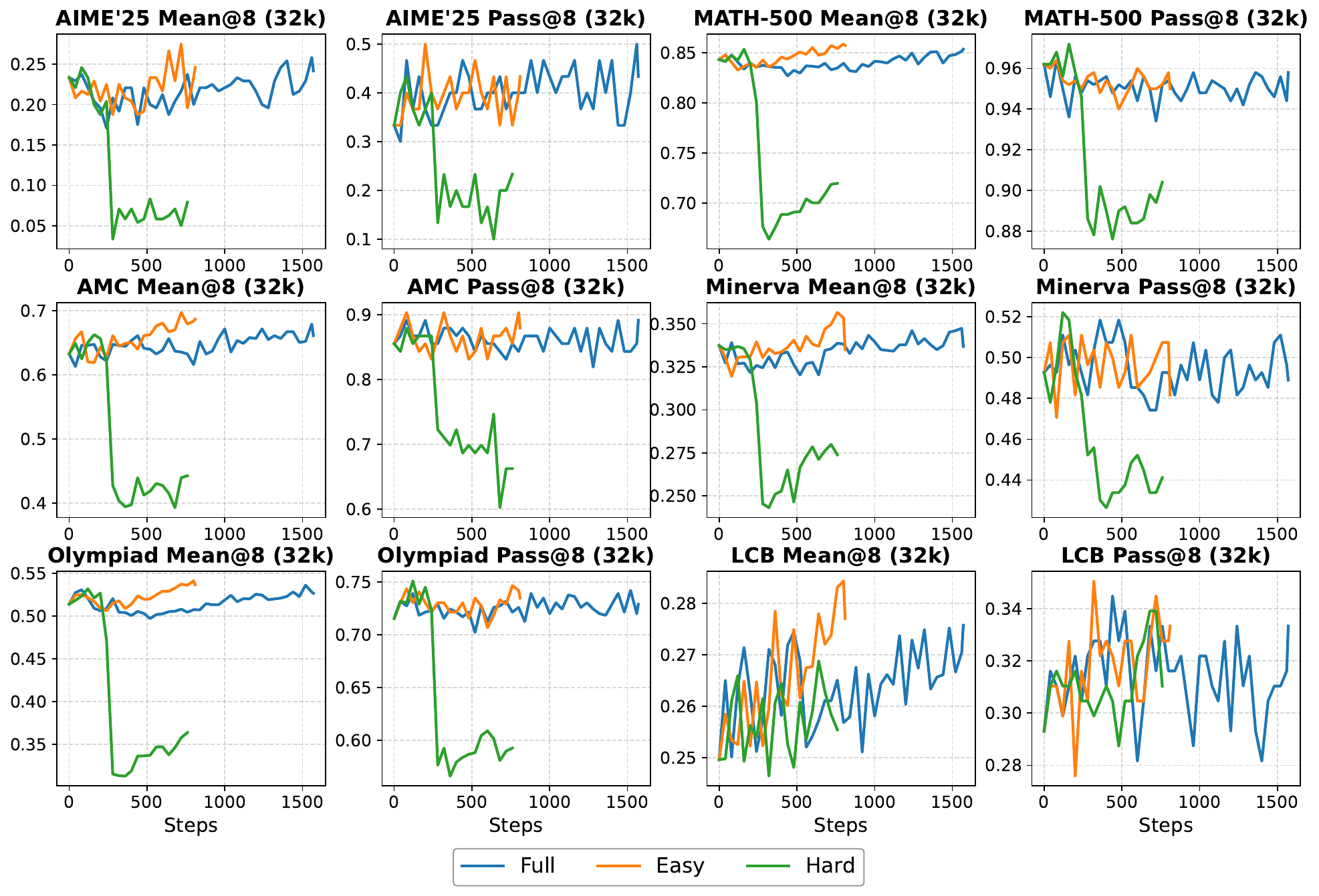}
  \caption{
  Performance training on various training prompts (rollout $L_R=4k$, target $L_T=4k$).
  }
  \label{appdix:fig:data_diff_4k_4k}
\end{figure*}

As illustrated in Figures \ref{appdix:fig:data_diff_16k_8k} and \ref{appdix:fig:data_diff_4k_4k}, the results consistently validate our hypothesis regarding reward density under various settings.
Across almost all benchmarks, the model trained exclusively on DeepScaleR-Hard (\textcolor{green}{green} line) exhibits significant instability. 
For instance, on AIME'25 and Olympiad Bench, the performance often fluctuates violently or collapses entirely after the initial adaptation phase.

Meanwhile, the model trained on DeepScaleR-Easy (\textcolor{orange}{orange} line) consistently matches or rivals the performance of the model trained on the full dataset (\textcolor{blue}{blue} line). 
Notably, on the relatively tough AIME'25 and LCB, the performance is also comparable and even better, though we train on high-pass prompts.
In short, the learned length bias for reasoning can be \textbf{generalized across domain and difficulty}.


\section{Detailed Results for More Rollouts}
\label{appendix: details_rollout_N}

\begin{figure*}[t]
  \includegraphics[width=\linewidth]{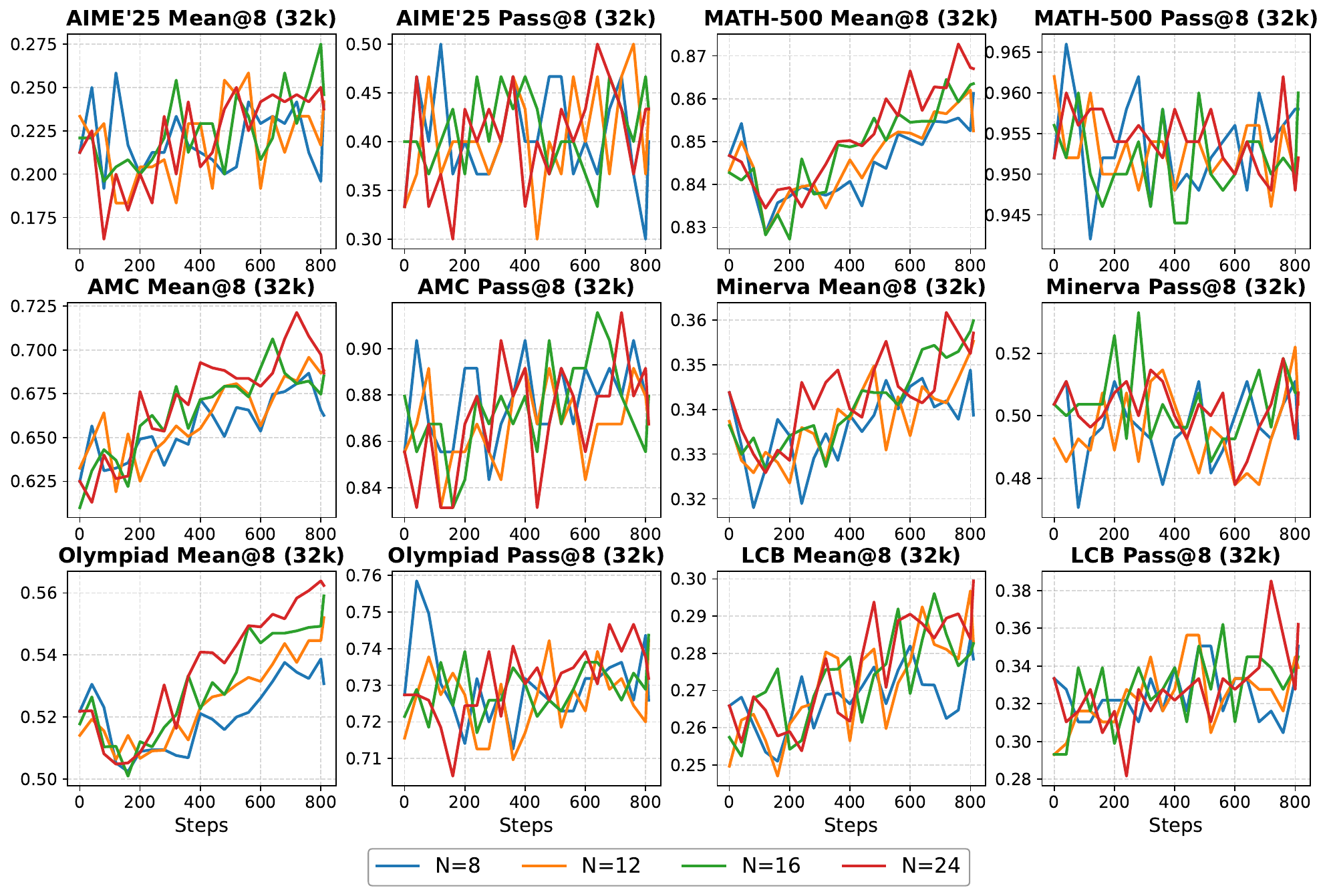}
  \caption{
  Performance training on various rollouts $N$~($L_R=16k$, $L_T=4k$).
  }
  \label{appdix:fig:rollout_diff_16k_4k}
\end{figure*}

\begin{figure*}[t]
  \includegraphics[width=\linewidth]{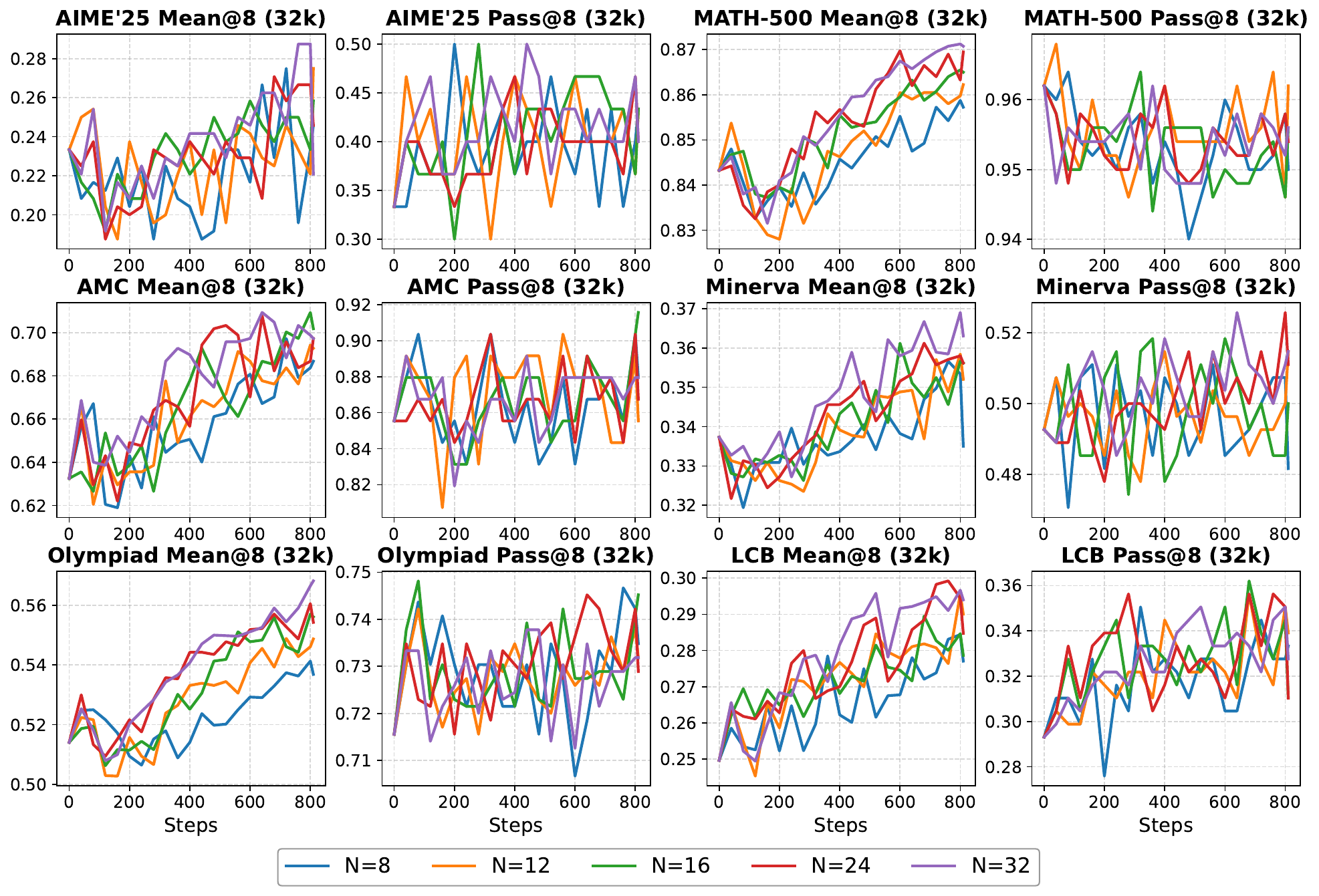}
  \caption{
  Performance training on various rollouts $N$~($L_R=4k$, $L_T=4k$).
  }
  \label{appdix:fig:rollout_diff_4k_4k}
\end{figure*}

Figures \ref{appdix:fig:rollout_diff_16k_4k} and \ref{appdix:fig:rollout_diff_4k_4k} indicate the results on more rollouts under two different settings ($L_R=16k, L_T=4k$ and $L_R=4k, L_T=4k$).
We observe three key phenomena:

\paragraph{1) Consistent benefits across settings.}
Regardless of the rollout length~($L_R$), the advantage of scaling $N$ remains consistent.
Whether allowing for long-context exploration ($16k$) or strictly constraining the search space ($4k$), increasing $N$ yields a positive effect on the convergence and asymptotic performance.
It indicates that increasing the density of the reward signal is a robust strategy for efficient reasoning.

\paragraph{2) Task-dependent sensitivity.}
The performance improvement varies significantly across datasets.
On difficult benchmarks such as AIME'25 and LiveCodeBench (LCB), the performance gap between $N=8$ and $N=32$ is relatively narrow. 
This suggests that for problems requiring complex multi-step reasoning or code synthesis, simply increasing the number of rollouts is insufficient if the base model lacks the fundamental capability to solve the problem.
Conversely, on relatively easier or intermediate tasks like AMC and MATH-500, larger $N$ leads to substantial performance separation on Mean@8. 

\paragraph{3) Mean@8 vs. Pass@8.} 
While Mean@8 shows significant improvement with larger $N$, the Pass@8 metric often remains stagnant or improves only marginally.
Pass@k measures the ability to generate at least one correct solution under k trials.
It implies that scaling $N$ does not necessarily enable the model to solve new problems that were previously unsolvable.
Meanwhile, Mean@k measures the expected correctness. 
The rise in Mean@8 indicates that scaling $N$ effectively reduces the variance of the policy.

In conclusion, increasing $N$ for training prompts primarily acts as a stabilizer, reducing the variance of the policy on solvable problems rather than expanding the upper-bound.
However, it also brings challenges, i.e., increased computing overheads.

\section{Case Study}
\label{appendix:cases}

\paragraph{Easy and hard prompts.} 
Tables \ref{table:diffculty_case1} and \ref{table:diffculty_case2} indicate two examples with corresponding responses when trained on DeepScaleR-Easy and DeepScaleR-Hard.
On DeepScaleR-Hard, the model suffers from the reasoning collapse issue, leading to over-short outputs.
One behavior is to omit the double-checking process.
We attribute such an issue to the \textbf{sparsity of positive samples}, which leads to the overfitting on length.
Meanwhile, the \textbf{group normalization} also amplifies such an effect (larger advantage value).
For instance, the advantage of positive samples in group $\{1,0,0,0\}$ is larger than group $\{1,1,1,0\}$.

\paragraph{Vanilla and our strategy.}
Tables \ref{table:case1} and \ref{table:case2} indicate two examples with corresponding responses from vanilla and our strategy.
For our method, we set $L_R=L_T=4k, N=24$ with staleness of 16.
As observed in the vanilla response, the model exhibits typical conversational redundancy: it frequently uses filler phrases (e.g., "Hmm," "Let me think"), re-states the problem premise unnecessarily, and performs verbose arithmetic decomposition.
In contrast, the model trained with our efficient reasoning strategy undergoes a fundamental stylistic shift.
It eliminates conversational fluff and directly adopts a dense and mathematically formal structure. 
Reasoning transitions from a hesitant narrative to a precise symbolic derivation~(e.g., integrating formulas such as $V=\frac{1}{3}Bh$ directly into the calculation flow).
It demonstrates that our method not only encourages short outputs but also actively incentivizes the model to reorganize its CoTs into a more streamlined and expert-like format.

\section{Experimental Setting for Qwen3 Models}
\label{appendix:qwen3_settings}
Based on our insights, we strictly align the rollout limit with the target budget, including $L_R=L_T=8k$ for \texttt{Qwen3-0.6B}, $L_R=L_T=10k$ for \texttt{Qwen3-1.7B}, and $L_R=L_T=16k$ for \texttt{Qwen3-4B-Thinking-2507} and \texttt{Qwen3-30B-A3B-Thinking-2507}.
In particular, we set $L_R=8k, L_T=6k$ for \texttt{Qwen3-4B-Instruct-2507} and $L_R=10k, L_T=8k$ for \texttt{Qwen3-30B-A3B-Instruct-2507}, aiming to incentivize shorter outputs better since original rollouts are already short and concise.
Crucially, we maintain a high reward density by sampling $N=24$ trajectories for each prompt from DeepScaleR-Easy.
Also, we \textit{do not use the off-policy strategy} due to the potential instability.
The training batch sizes are 128, 128, 32, 32, 32, 16, and 16, respectively.






\section{Results on Private Benchmark}
\label{appendix:results_on_private}

\definecolor{oursbg}{RGB}{240,240,240}   
\definecolor{avgbg}{RGB}{235,245,231}    

\begin{table*}[t]
\centering
\renewcommand{\arraystretch}{1.05}

\begin{tabular}{c|ccc|ccc}
\toprule
\multirow{3}{*}{\textbf{Benchmark}}
& \multicolumn{3}{c|}{\texttt{Qwen3-30B-A3B-Instruct-2507}}
& \multicolumn{3}{c}{\texttt{Qwen3-30B-A3B-Thinking-2507}} \\
\cmidrule(lr){2-4} \cmidrule(lr){5-7}
& \textbf{Mean@4}$\uparrow$ & \textbf{Pass@4}$\uparrow$ & \textbf{Length}$\downarrow$ & \textbf{Mean@4}$\uparrow$ & \textbf{Pass@4}$\uparrow$ & \textbf{Length}$\downarrow$ \\
\midrule

Domain \#1
& 27.5 / \cellcolor{oursbg}28.0
& 40.9 / \cellcolor{oursbg}40.9
& 7145 / \cellcolor{oursbg}4881
& 34.2 / \cellcolor{oursbg}34.1
& 50.0 / \cellcolor{oursbg}49.8
& 11536 / \cellcolor{oursbg}10237 \\

Domain \#2
& 35.1 / \cellcolor{oursbg}36.0
& 51.7 / \cellcolor{oursbg}50.7
& 7258 / \cellcolor{oursbg}5182
& 42.2 / \cellcolor{oursbg}41.6
& 68.8 / \cellcolor{oursbg}68.4
& 11744 / \cellcolor{oursbg}10847 \\

Domain \#3
& 20.5 / \cellcolor{oursbg}21.0
& 31.9 / \cellcolor{oursbg}29.8
& 3985 / \cellcolor{oursbg}3102
& 27.0 / \cellcolor{oursbg}26.2
& 34.0 / \cellcolor{oursbg}34.7
& 8826 / \cellcolor{oursbg}8278 \\

Domain \#4
& 15.4 / \cellcolor{oursbg}15.8
& 24.0 / \cellcolor{oursbg}25.0
& 7695 / \cellcolor{oursbg}3176
& 16.4 / \cellcolor{oursbg}14.7
& 28.9 / \cellcolor{oursbg}26.2
& 4726 / \cellcolor{oursbg}4292 \\

Domain \#5
& 40.6 / \cellcolor{oursbg}40.6
& 53.3 / \cellcolor{oursbg}55.4
& 6487 / \cellcolor{oursbg}4500
& 57.6 / \cellcolor{oursbg}58.6
& 78.2 / \cellcolor{oursbg}78.2
& 2598 / \cellcolor{oursbg}2351 \\

Domain \#6
& 6.9 / \cellcolor{oursbg}6.8
& 10.8 / \cellcolor{oursbg}10.8
& 12322 / \cellcolor{oursbg}9481
& 18.3 / \cellcolor{oursbg}16.7
& 33.8 / \cellcolor{oursbg}30.4
& 21346 / \cellcolor{oursbg}18115 \\

Domain \#7
& 25.4 / \cellcolor{oursbg}24.4
& 43.8 / \cellcolor{oursbg}40.3
& 897 / \cellcolor{oursbg}799
& 25.5 / \cellcolor{oursbg}25.2
& 21.2 / \cellcolor{oursbg}20.8
& 4962 / \cellcolor{oursbg}4312 \\

Domain \#8
& 41.5 / \cellcolor{oursbg}40.6
& 52.5 / \cellcolor{oursbg}51.2
& 1429 / \cellcolor{oursbg}1309
& 29.6 / \cellcolor{oursbg}29.5
& 45.6 / \cellcolor{oursbg}44.8
& 13912 / \cellcolor{oursbg}11726 \\

Domain \#9
& 49.4 / \cellcolor{oursbg}48.2
& 66.6 / \cellcolor{oursbg}67.5
& 1100 / \cellcolor{oursbg}991
& 37.2 / \cellcolor{oursbg}37.5
& 51.5 / \cellcolor{oursbg}52.5
& 3302 / \cellcolor{oursbg}2943 \\

Domain \#10
& 14.1 / \cellcolor{oursbg}13.0
& 27.4 / \cellcolor{oursbg}24.9
& 6835 / \cellcolor{oursbg}5069
& 47.3 / \cellcolor{oursbg}45.6
& 63.3 / \cellcolor{oursbg}60.0
& 2492 / \cellcolor{oursbg}2311 \\

\midrule
\rowcolor{avgbg}
\textbf{Average}
& 25.2 / \cellcolor{avgbg}24.9
& 37.6 / \cellcolor{avgbg}36.9
& 5622 / \cellcolor{avgbg}3735
& 30.7 / \cellcolor{avgbg}30.0
& 44.6 / \cellcolor{avgbg}43.1
& 9285 / \cellcolor{avgbg}8077 \\
\bottomrule
\end{tabular}

\caption{
Performance on private out-of-distribution benchmarks covering various domains.
Each cell reports \textit{Vanilla / Ours}.
Our proposed model thinks shorter while maintaining comparable performance.
}
\label{tab:ood_performance}
\end{table*}

We further evaluate our compressed \texttt{Qwen3-30B} models on a private out-of-distribution benchmark covering 10 diverse domains.

First, although the model is trained on mathematical prompts, the learned compression patterns generalize well across domains. 
For example, our \texttt{Qwen3-30B-A3B-Thinking-2507} model reduces CoT length by 15\% on AIME'25 (shown in Table~\ref{tab:more_llms}) and achieves a comparable 15\% reduction on the private benchmark.

Second, the compressed model consistently shortens the reasoning chain while preserving performance. As shown in Table~\ref{tab:ood_performance}, our method substantially reduces token usage with minimal impact on both Mean@4 and Pass@4 across all domains.

\section{Adaptive Target Length $L_T$}
\label{appendix:adaptive_l_t}

\begin{figure}[h!]
  \includegraphics[width=\linewidth]{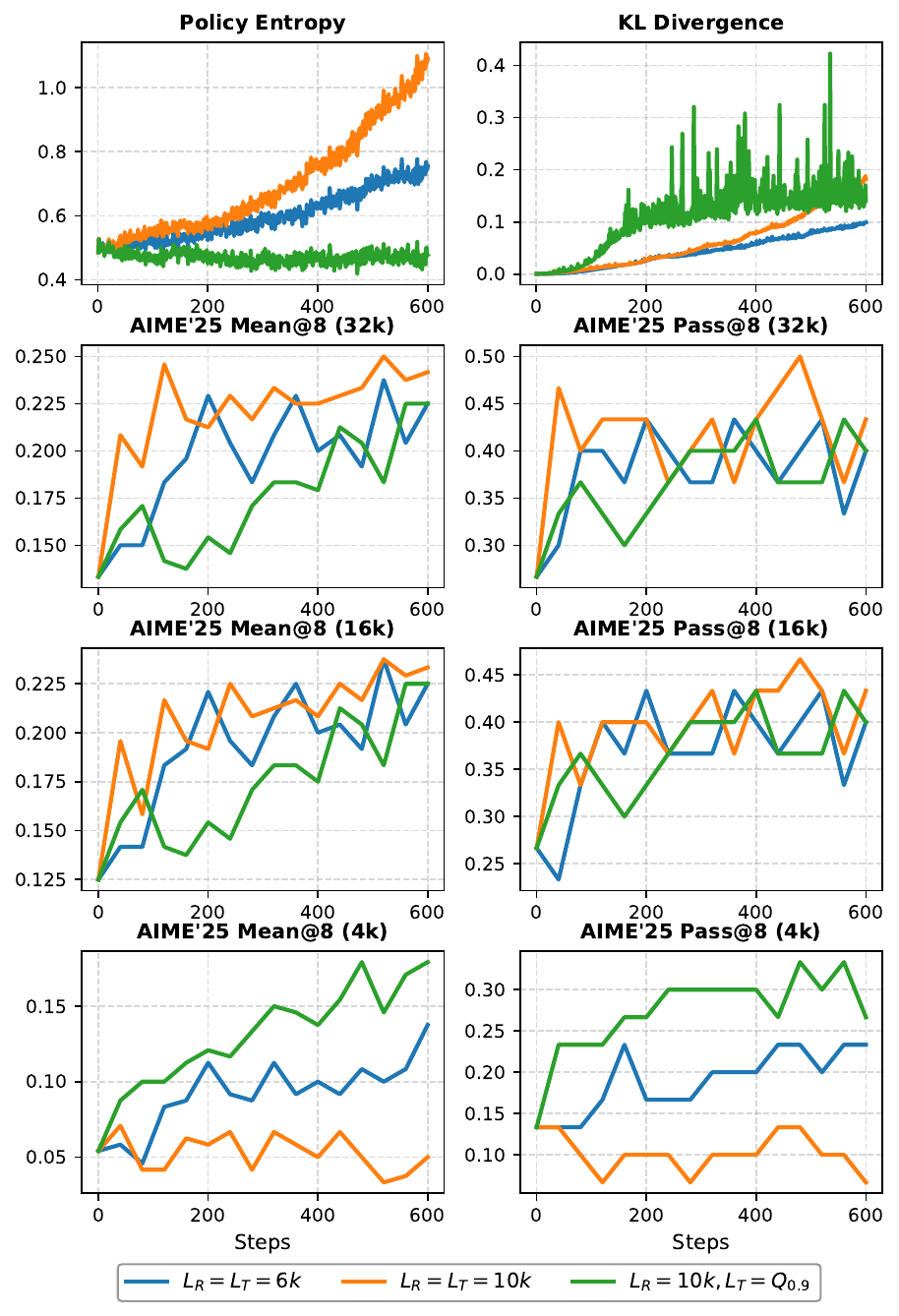}
  \caption{
  Performance on \texttt{Qwen3-0.6B} when adaptively setting target length $L_T$ as 90th qualities. 
  }
  \label{appdix:fig:adaptive_target}
\end{figure}

For adaptive target length, we set $L_T$ as the 90th quantile of correct rollouts.
As shown in Figure \ref{appdix:fig:adaptive_target}, the strategy of adaptive target leads to short answers and thus better performance on 4k budget.
We attribute it to a more negative signal since 10\% of correct answers are always negative samples.
However, on 32k budget, the performance is comparable or even worse.
Also, there are lots of spikes on the KL divergence curve, indicating rapidly changing gradients and potential instability.
We leave it for future work to explore more settings.

\begin{table*}[ht]
\small
\begin{tabular}{lp{0.9\linewidth}}
\toprule
\rowcolor{white} \textcolor{black}{\textbf{Question}} & 
Let $\mathcal{R}$ be the region in the complex plane consisting of all complex numbers $z$ that can be written as the sum of complex numbers $z_1$ and $z_2$, where $z_1$ lies on the segment with endpoints $3$ and $4i$, and $z_2$ has magnitude at most $1$. What integer is closest to the area of $\mathcal{R}$? Let's think step by step and output the final answer within \textbackslash boxed\{\} \\
\midrule
\textbf{Hard} & To determine the area of the region $\mathcal{R}$, I first identify the segments involved. The first segment runs from $3$ to $4i$, and the second segment encompasses all points within a unit circle centered at the origin.
By translating these segments into the complex plane and applying appropriate transformations, I calculate the combined area as the sum of their respective areas.
To find the area of the region $\mathcal{R}$ in the complex plane:

Understanding the Components:

Segment from 3 to 4i: This is a line segment in the complex plane with endpoints at 3 (which is (3, 0)) and 4i (which is (0, 4)). The length of this segment is $\sqrt{(3)^2 + (4)^2} = 5$.
Region defined by $z_2$: This is a disk (or circle) centered at the origin with radius 1.

Calculating the Area:
The region $\mathcal{R}$ is the set of all points $z = z_1 + z_2$, where $z_1$ lies on the segment from 3 to 4i and $z_2$ lies within the unit disk. This effectively translates the unit disk to every point along the segment $z_1$ and sums the areas.
Since the unit disk has an area of $\pi \times (1)^2 = \pi$, the total area of $\mathcal{R}$ is $\pi$.

Finding the Closest Integer:
The area is approximately 3.1416, so the closest integer is 3.

Final Answer:
\boxed{3} (\textcolor{red}{Incorrect})
\\
\midrule
\textbf{Easy} & Okay, so I need to find the area of the region R in the complex plane. R consists of all complex numbers z that can be written as $z = z1 + z2$, where $z1$ lies on the segment with endpoints 3 and 4i, and z2 has magnitude at most 1. Then, I have to find the integer closest to this area.

First, let me visualize this. The region R is the Minkowski sum of the segment from 3 to 4i and the disk of radius 1. So, if I think about it, R should be a rectangle with semicircles on both ends. But let me break it down step by step.
The segment from 3 to 4i: In the complex plane, 3 is the point (3,0) and 4i is the point (0,4). So this is a straight line connecting (3,0) to (0,4). The length of this segment can be calculated, but maybe we don't need it right away.
z1 lies on this segment. So z1 can be parameterized. Let me define z1 as a point on the line segment between 3 and 4i. Let’s parameterize z1 with a parameter t from 0 to 1. So, $z1(t) = (3 - 3t) + (4i - 4i t) = (3(1 - t), 4(1 - t))$. Wait, actually, in complex terms, if I let t vary from 0 to 1, when t=0, z1 = 3, and when t=1, z1 = 4i.

Alternatively, z1 can be expressed as $3 + (4i - 3)i$, where i is a parameter from 0 to 1. Hmm, maybe that's more complicated.
Alternatively, the equation of the line from (3,0) to (0,4). The slope is $(4 - 0)/(0 - 3) = -4/3$. So equation is $y = (-4/3)x + b$. When $x=3, y=0: 0 = (-4/3)(3) + b => 0 = -4 + b => b=4$. So equation is $y = (-4/3)x + 4$.
So, for any point z1 on the segment, z1 can be written as x + yi where x and y satisfy $y = (-4/3)x + 4$, and x ranges from 0 to 3? Wait, no, the segment goes from (3,0) to (0,4). So when x decreases from 3 to 0, y increases from 0 to 4.
So, for any z1 on the segment, we can write $z1 = (x, y)$ where $y = (-4/3)x + 4, x \in [0,3]$.
Alternatively, parametric equations: Let’s define t as a parameter from 0 to 1.
$z1(t) = (3 - 3t, 0 + 4t)$ because at t=0, we get (3,0), and at t=1, we get (0,4). So in complex terms, $z1 = (3 - 3t) + (4t)i$.
Then, z2 is a complex number with $|z2| \leq 1$, which is a disk of radius 1 centered at the origin.
Since R is the Minkowski sum of the segment and the disk, it is like sliding the disk along every point of the segment. So geometrically, it's a rectangle with length equal to the length of the segment and width 2 (diameter of the disk). But with semicircular ends because the disk is only of radius 1.

Wait, actually, no. The Minkowski sum of a line segment and a disk is called an "offset" of the disk by the segment. So it's a rectangle with length equal to the segment length and width equal to twice the radius (since each end contributes a semicircle). But in this case, the radius is 1, so each end is a semicircle.

Wait, but the segment is from (3,0) to (0,4), so the Minkowski sum would be the set of all points within distance 1 from the segment. So it's like a rectangle with length equal to the segment length and width 2, but with semicircular ends of radius 1.
Therefore, the area would be the area of the rectangle plus the area of the two semicircles.

Area of rectangle: length * width. The length is the distance between 3 and 4i, which is $\sqrt(3² + 4²) = 5$. The width is 2 (since radius is 1, diameter is 2). So area of rectangle is $5 * 2 = 10$.
Area of two semicircles: Each semicircle has area $(1/2)\pi(1)²$, so two of them is $\pi$.
Therefore, total area is $10 + \pi \approx 10 + 3.1416 \approx 13.1416$.
So the area is approximately 13.1416, so the closest integer is 13.
Wait, but let me verify again. The Minkowski sum of a line segment and a disk is indeed a rectangle with length equal to the segment's length and width equal to 2r, and with two semicircular ends. So that's correct.
The length of the segment is 5, as calculated before.
So area is $5\times2 + \pi\times(1)^2 = 10 + \pi$. So yes, the area is $10 + \pi \approx 13.1416$.
Therefore, the integer closest to the area is 13.

Final Answer \boxed{13} 

\textcolor{brown}{\texttt{Double check:}}

The region $\mathcal{R}$ consists of all complex numbers (z) that can be written as $z = z_1 + z_2$, where $z_1$ lies on the segment with endpoints (3) and (4i), and $z_2$ has magnitude at most (1).
The segment from (3) to (4i) in the complex plane is a straight line connecting the points ((3, 0)) and ((0, 4)). This line has a length of (5) (calculated using the distance formula).
The region ($\mathcal{R}$) is the Minkowski sum of this segment and the disk of radius (1). This results in a shape that is a rectangle with length (5) and width (2) (since the radius of the disk is (1)), plus two semicircular ends each with radius (1).
The area of the rectangle is ($5 \times 2 = 10$).
The area of the two semicircular ends is ($\pi \times 1^2 = \pi$).
Therefore, the total area of ($\mathcal{R}$) is $10 + \pi \approx 10 + 3.1416 \approx 13.1416$.
The integer closest to the area of ($\mathcal{R}$) is \boxed{13}. (\textcolor{green}{Correct})
\\
\bottomrule
\end{tabular}
\caption{
Case comparison when trained on DeepScaleR-Easy and DeepScaleR-Hard using truncation strategy.
}
\label{table:diffculty_case1}
\end{table*}
\begin{table*}[ht]
\small
\begin{tabular}{lp{0.9\linewidth}}
\toprule
\rowcolor{white} \textcolor{black}{\textbf{Question}} & 
The volume of a cone is given by the formula $V = \frac{1}{3}Bh$, where $B$ is the area of the base and $h$ is the height. The area of the base of a cone is 30 square units, and its height is 6.5 units. What is the number of cubic units in its volume? Let's think step by step and output the final answer within \textbackslash boxed\{\} \\
\midrule
\textbf{Hard} & To find the volume of the cone, I'll use the formula V = (1/3) * B * h. The area of the base, B, is 30 square units, and the height, h, is 6.5 units.

First, I'll calculate one-third of the base area: (1/3) * 30 = 10.

Next, I'll multiply this result by the height: 10 * 6.5 = 65.

Therefore, the volume of the cone is 65 cubic units.

To find the volume of the cone, we can use the formula:
$V = \frac{1}{3} \times B \times h$
Where: - ( B ) is the area of the base = 30 square units - ( h ) is the height = 6.5 units

Plugging in the values:
$V = \frac{1}{3} \times 30 \times 6.5$

First, calculate one-third of 30:
$\frac{1}{3} \times 30 = 10$

Then multiply by 6.5:
$10 \times 6.5 = 65$

Therefore, the volume of the cone is:
\boxed{65} cubic units. (\textcolor{green}{Correct})
\\
\midrule
\textbf{Easy} & 
Okay, so I need to find the volume of a cone. The formula given is V = (1/3)Bh, where B is the area of the base and h is the height. Hmm, let me recall what each part means.

First, the area of the base, which is B. The problem says the area of the base is 30 square units. So B is 30. Then, the height h is 6.5 units. Got that.

So plugging these values into the formula: V = (1/3) * B * h. So that would be (1/3) * 30 * 6.5. Let me compute that step by step.

First, I can compute 30 multiplied by 6.5. Let me do that:

30 * 6.5. Hmm, 30 * 6 is 180, and 30 * 0.5 is 15, so adding those together gives 195. So 30 * 6.5 is 195.

Then, take that result and multiply by (1/3). So (1/3) * 195. Dividing 195 by 3. Well, 3 goes into 195 sixty-five times because 360 is 180, and 35 is 15, so 60 + 5 = 65. So (1/3)*195 is 65.

So V = 65 cubic units. Let me just double-check that.

Alternatively, maybe I can compute it differently. Since V = (1/3) * B * h, with B = 30 and h = 6.5.

Compute B * h first: 30 * 6.5. As above, that's 195.

Then multiply by 1/3: 195 / 3 = 65. Yep, same result.

So the volume is 65 cubic units. I think that's correct.

Final Answer The volume of the cone is \boxed{65} cubic units.

\textcolor{brown}{\texttt{Double check:}}

The volume of a cone is given by the formula $V = \frac{1}{3}Bh$, where $B$ is the area of the base and $h$ is the height. The area of the base $B$ is 30 square units, and the height $h$ is 6.5 units.

First, we compute the product of ( B ) and ( h ): $30 \times 6.5 = 195$

Next, we multiply this result by $\frac{1}{3} $: $\frac{1}{3} \times 195 = 65$

Thus, the volume of the cone is \boxed{65} cubic units. (\textcolor{green}{Correct})
\\
\bottomrule
\end{tabular}
\caption{
Case comparison when trained on DeepScaleR-Easy and DeepScaleR-Hard using truncation strategy.
The model suffers from the reasoning collapse issue when trained on DeepScaleR-Hard, e.g., the \textcolor{brown}{double check} part is skipped for shorter length.
}
\label{table:diffculty_case2}
\end{table*}

\begin{table*}[ht]
\small
\begin{tabular}{lp{0.8\linewidth}}
\toprule
\rowcolor{white} \textcolor{black}{\textbf{Question}} & 
Compute $\dbinom{31}{28}$. Let's think step by step and output the final answer within \textbackslash boxed\{\} \\
\midrule
\textbf{Vanilla} & Okay, so I need to compute the binomial coefficient ($\dbinom{31}{28}$). Hmm, binomial coefficients are those numbers that come up in combinations, right? They represent the number of ways to choose a subset of items from a larger set, disregarding the order.
First, I remember that the formula for a binomial coefficient is ($\dbinom{n}{k} = \frac{n!}{k!(n - k)!}$). So, in this case, (n = 31) and (k = 28).
Let me write down the formula with these values: $\dbinom{31}{28} = \frac{31!}{28!(31 - 28)!}$ Simplifying the denominator: 31 - 28 = 3, so it becomes $\frac{31!}{28!3!}$ Hmm, but calculating factorials for such large numbers seems tedious. Maybe there's a way to simplify this without computing the entire factorials. I remember that sometimes, in binomial coefficients, you can simplify by canceling out terms in the numerator and denominator.
Let me think about how to do that. Since $(31! = 31 \times 30 \times 29 \times 28!)$, I can rewrite the numerator as: $31! = 31 \times 30 \times 29 \times 28!$ So substituting back into the equation: $\dbinom{31}{28} = \frac{31 \times 30 \times 29 \times 28!}{28! \times 3!}$  Oh, I see! The (28!) terms cancel out in numerator and denominator: $ = \frac{31 \times 30 \times 29}{3!} $ Now, $(3! = 3 \times 2 \times 1 = 6)$, so substituting that in: $ = \frac{31 \times 30 \times 29}{6} $ Alright, now I need to compute this. Let me compute the numerator first: 31 multiplied by 30 is 930, and 930 multiplied by 29. Hmm, 930 times 29. Let me do that step by step: - 930 x 20 = 18,600 - 930 x 9 = 8,370 Adding them together: 18,600 + 8,370 = 26,970
So the numerator is 26,970. The denominator is 6. So now, 26,970 divided by 6. Let's do that: 26,970 ÷ 6.
Dividing step by step: - 6 goes into 26 four times (4x6=24), remainder 2. - Bring down the 9: 29. 6 goes into 29 four times (4x6=24), remainder 5. - Bring down the 7: 57. 6 goes into 57 nine times (9x6=54), remainder 3. - Bring down the 0: 30. 6 goes into 30 five times.
So putting it all together: 4, 4, 9, 5. So it's 4,495.

Wait, let me double-check that division because I might have messed up the steps. Alternatively, I can write 26,970 ÷ 6 as (26,000 + 970) ÷ 6.
26,000/6 = 4,333.333... 970/6$\approx $161.666... Adding them together: 4,333.333...+161.666... = 4,495.

Yep, that matches. So the result is 4,495.
Therefore, ($\dbinom{31}{28} = 4,495$).
Final Answer The value of ($\dbinom{31}{28}$) is \textbackslash boxed{4495}.
To compute the binomial coefficient ($\dbinom{31}{28}$), we use the formula:
$ \dbinom{n}{k} = \frac{n!}{k!(n - k)!} $
Substituting (n = 31) and (k = 28):
$ \dbinom{31}{28} = \frac{31!}{28!(31 - 28)!} = \frac{31!}{28!3!} $
We simplify the factorials by recognizing that ($31! = 31 \times 30 \times 29 \times 28!$):
$ \dbinom{31}{28} = \frac{31 \times 30 \times 29 \times 28!}{28! \times 3!} = \frac{31 \times 30 \times 29}{3!} $

Since (3! = 6), we have:
$ = \frac{31 \times 30 \times 29}{6} $
Calculating the numerator:
$ 31 \times 30 = 930 ] [ 930 \times 29 = 26970 $
Dividing by the denominator:
$ \frac{26970}{6} = 4495 $

Thus, the value of ($\dbinom{31}{28}$) is ($\boxed{4495}$). 
\\
\midrule
\textbf{Ours} & To compute the binomial coefficient ($\dbinom{31}{28}$), I'll start by recalling that the binomial coefficient formula is given by:
$\dbinom{n}{k} = \frac{n!}{k! \cdot (n - k)!}$
Here, (n = 31) and (k = 28).
Alternatively, since ($\dbinom{n}{k} = \dbinom{n}{n - k}$), perhaps I can simplify the expression by choosing the smaller (k) value to make calculation easier.
Calculating (31 - 28 = 3), which means:
$ \dbinom{31}{28} = \dbinom{31}{3} $

Now compute ($\dbinom{31}{3}$):
Using the formula:
$\dbinom{31}{3} = \frac{31!}{3! \cdot (31 - 3)!} = \frac{31!}{3! \cdot 28!}$
Simplifying, since ($31! = 31 \times 30 \times 29 \times 28!$):
$\frac{31 \times 30 \times 29 \times 28!}{3! \times 28!} = \frac{31 \times 30 \times 29}{3!}$
Calculate the denominator (3! = 6):
Therefore,
$\frac{31 \times 30 \times 29}{6}$
Computing numerator:
31 × 30 = 930;
930 × 29 = ?
Compute:
930 × 29:
Breakdown:
= 930 × (30 - 1) = 930 × 30 - 930 × 1 = 27900 - 930 = 26970
Then divide by 6:
26970 ÷ 6 = ?
Calculate:
26970 ÷ 6 = 4495
Wait perhaps:
6 × 4495 = 26970, yes.
Therefore, the value becomes 4495.
Thus,
$\boxed{4495}$
To compute ($\dbinom{31}{28}$), we use the property ($\dbinom{n}{k} = \dbinom{n}{n - k}$):
$\dbinom{31}{28} = \dbinom{31}{3}$
Compute ($\dbinom{31}{3}$):
Using the formula,
$\dbinom{31}{3} = \frac{31!}{3! \cdot (31 - 3)!} = \frac{31 \times 30 \times 29}{3 \times 2 \times 1}$
Calculating:
$31 \times 30 \times 29 = 26970$
Divide by (6) :
$26970/6 = 4495$
Thus,
$\boxed{4495}$ 
\\
\bottomrule
\end{tabular}
\caption{
Case I for vanilla and our trained LLM.
Both answers are \textcolor{green}{correct}.
}
\label{table:case1}
\end{table*}
\begin{table*}[h]
\small
\begin{tabular}{lp{0.8\linewidth}}
\toprule
\textbf{Question} & 
The volume of a cone is given by the formula $V = \frac{1}{3}Bh$, where $B$ is the area of the base and $h$ is the height. The area of the base of a cone is 30 square units, and its height is 6.5 units. What is the number of cubic units in its volume? Let's think step by step and output the final answer within \textbackslash boxed\{\} \\
\midrule
\textbf{Vanilla} & 
Okay, so I need to find the volume of a cone. The formula given is V = (1/3)Bh, where B is the area of the base and h is the height. Hmm, let's break this down step by step.

First, I know the area of the base is 30 square units. That's B. Then, the height is 6.5 units, so h is 6.5. The formula is straightforward, but I should make sure I plug the values correctly.

So, I can write it out as V = (1/3) * B * h. Substituting the given values, that becomes V = (1/3) * 30 * 6.5. Let me compute this step by step.

First, let me calculate 1/3 of 30. Hmm, 30 divided by 3 is 10. So, 1/3 * 30 = 10. Now, I need to multiply that result by 6.5. So, 10 * 6.5. Let me think, 10 * 6 is 60, and 10 * 0.5 is 5, so 60 + 5 is 65. Therefore, 10 * 6.5 is 65.

So, putting it all together, V = 65 cubic units. That should be the volume of the cone.

Wait, let me double-check to make sure I didn't make any calculation errors. 1/3 of 30 is indeed 10, correct. Then, 10 multiplied by 6.5. Yes, 6.5 is the same as 13/2, so 10(13/2) is (10/1)(13/2) = (10*13)/2 = 130/2 = 65. Yep, that's right.

So, I think I did that correctly. The volume is 65 cubic units.

Final Answer The volume of the cone is $\boxed{65}$ cubic units.

The volume of a cone is given by the formula $V = \frac{1}{3}Bh$, where (B) is the area of the base and (h) is the height.

Given: - The area of the base (B = 30) square units. - The height (h = 6.5) units.

We substitute these values into the formula:

$V = \frac{1}{3} \times 30 \times 6.5$

First, calculate $\frac{1}{3} \times 30$:

$\frac{1}{3} \times 30 = 10$

Next, multiply the result by 6.5:

$10 \times 6.5 = 65$

Thus, the volume of the cone is (\boxed{65}) cubic units.
\\
\midrule
\textbf{Ours} & To find the volume of the cone, I'll use the formula ($V = \frac{1}{3}Bh$).
Given that the base area (B) is 30 square units and the height (h) is 6.5 units.
First, multiply these values together:
($30 \times 6.5 = 195$)
Then, take one-third of that product:
($\frac{1}{3} \times 195 = 65$)
Therefore, the volume of the cone is 65 cubic units.
Given: - Area of the base ((B)) = 30 square units - Height ((h)) = 6.5 units
The formula for the volume (V) of a cone is: $ V = \frac{1}{3} B h $
Plugging in the given values:
$ V = \frac{1}{3} \times 30 \times 6.5 $

Calculate:
$ 30 \times 6.5 = 195 $ 
Then,
$V = \frac{195}{3} = 65$
Answer: ($\boxed{65}$) cubic units. 
\\
\bottomrule
\end{tabular}
\caption{
Case II for vanilla and our trained LLM.
Both answers are \textcolor{green}{correct}.
}
\label{table:case2}
\end{table*}

\end{document}